\documentclass[letterpaper]{article} 

\usepackage{aaai2026}  

\usepackage{times}  
\usepackage{helvet}  
\usepackage{courier}  
\usepackage[hyphens]{url}  
\usepackage{graphicx} 
\usepackage{natbib}  
\usepackage{caption} 
\usepackage{algorithm}
\usepackage{algorithmic}

\usepackage{newfloat}
\usepackage{listings}
\DeclareCaptionStyle{ruled}{labelfont=normalfont,labelsep=colon,strut=off} 
\floatstyle{ruled}
\newfloat{listing}{tb}{lst}{}
\floatname{listing}{Listing}
\usepackage[utf8]{inputenc} 
\usepackage[T1]{fontenc}    
\usepackage{url}            
\usepackage{booktabs}       
\usepackage{amsfonts}       
\usepackage{nicefrac}       
\usepackage{microtype}      
\usepackage{xcolor}         
\usepackage{threeparttable} 

\usepackage{amsmath}
\usepackage{amssymb}
\usepackage{mathtools}
\usepackage{amsthm}

\usepackage[capitalize,noabbrev]{cleveref}
\usepackage{yhmath}

\usepackage{multirow}
\usepackage{soul}
\usepackage{mathtools}
\usepackage{amsmath,amsfonts,amsthm}
\usepackage[algo2e,linesnumbered,ruled,vlined]{algorithm2e}
\usepackage{float}
\usepackage{lipsum}
\usepackage{textcomp}
\usepackage{xcolor}
\usepackage{soul}
\usepackage{placeins}
\usepackage{hhline}
\usepackage{tabularx}
\usepackage{enumitem}
\usepackage{booktabs}
\usepackage{adjustbox}
\usepackage{array}
\usepackage{colortbl}
\usepackage{subcaption}

\graphicspath{{figures/}}

\theoremstyle{plain}

\theoremstyle{definition}

\DeclareMathOperator*{\argmin}{arg\,min}

\usepackage{pifont}
\usepackage{bbding}
\newcommand{\xmark}{\ding{55}}

\title{$D^2 Prune:$ Sparsifying Large Language Models via Dual Taylor Expansion and Attention Distribution Awareness}
\title{$D^2 Prune:$ Sparsifying Large Language Models via Dual Taylor Expansion and Attention Distribution Awareness}

\author{
    Lang Xiong\textsuperscript{\rm 1}\equalcontrib,
    Ning Liu\textsuperscript{\rm 3}\equalcontrib,
    Ao Ren\textsuperscript{\rm 1}\thanks{Corresponding authors},
    Yuheng Bai\textsuperscript{\rm 1},
    Haining Fang\textsuperscript{\rm 1},
    Binyan Zhang\textsuperscript{\rm 1},
    Zhe Jiang\textsuperscript{\rm 1},
    Yujuan Tan\textsuperscript{\rm 2}\footnotemark[\value{footnote}],
    Duo Liu\textsuperscript{\rm 1}
}

\affiliations{
    \textsuperscript{\rm 1}Chongqing University, China\\
    \textsuperscript{\rm 2}National University of Defense Technology, China\\
    \textsuperscript{\rm 3}Beijing Innovation Center of Humanoid Robotics, China\\
    [3pt]
    langxiong@stu.cqu.edu.cn,\,
    ningliu1220@gmail.com,\,
    \{ren.ao, yuheng, haining.fang, zby, jiangzhe\}@cqu.edu.cn,\,
    tanyujuan@nudt.edu.cn,\,
    liuduo@cqu.edu.cn
}
\usepackage{bibentry}

\begin{document}

\maketitle

\begin{abstract}
Large language models (LLMs) face significant deployment challenges due to their massive computational demands. 
While pruning offers a promising compression solution, existing methods suffer from two critical limitations:
(1) They neglect activation distribution shifts between calibration data and test data, resulting in inaccurate error estimations;
(2) Overlooking the long-tail distribution characteristics of activations in the attention module. 
To address these limitations, this paper proposes a novel pruning method, $D^2Prune$. 
First, we propose a dual Taylor expansion-based method that jointly models weight and activation perturbations for precise error estimation, leading to precise pruning mask selection and weight updating and facilitating error minimization during pruning.
Second, we propose an attention-aware dynamic update strategy that preserves the long-tail attention pattern by jointly minimizing the KL divergence of attention distributions and the reconstruction error.
%
Extensive experiments show that $D^2Prune$ consistently outperforms SOTA meth
ods across various LLMs (e.g., OPT-125M, LLaMA2/3, Qwen3). Moreover, the dynamic attention update mechanism also generalizes well to ViT-based vision models like DeiT, achieving superior accuracy on ImageNet-1K.
\end{abstract}

\begin{links}
    \link{Code \& Full Version}{https://github.com/cquxl/D2Prune/}
\end{links}

\section{Introduction}
\label{sec: introduction}
In recent years, Large Language Models (LLMs)~\cite{guo2025deepseek, dubey2024llama, openai2023gpt4} have revolutionized the field of Natural Language Processing (NLP), excelling in complex tasks such as causal reasoning and text generation \cite{chi2024unveiling, wang2024causalbench, mo2024large, li2024pre}.
However, their exceptional performance comes at the cost of substantial memory and computational demands, posing significant challenges for deployment on resource-constrained devices \cite{li2024discovering, kim2024shortened}.
Extensive efforts have been made in model pruning techniques, 
aiming at shrinking the network size by removing redundant weights~\cite{bai2024sparsellm, dong2024pruner, sun2023simple, frantar2023sparsegpt}. 
By introducing sparsity, pruning enhances both memory and computational efficiency and has proven effective in LLMs.
Nonetheless, traditional training-based model pruning generally requires substantial computational resources due to processes such as retraining~\cite{liu2018rethinking, blalock2020state}, training from scratch~\cite{hoang2023revisiting, sreenivasan2022rare}, or extensive iterative pruning~\cite{chijiwa2021pruning, tanaka2020pruning}. 
These training-based methods are costly, especially for current large language models (LLMs). 
\begin{figure*}[t]
\centering
\subfloat[Upstream activation distribution]{
        \includegraphics[width=0.325\linewidth]{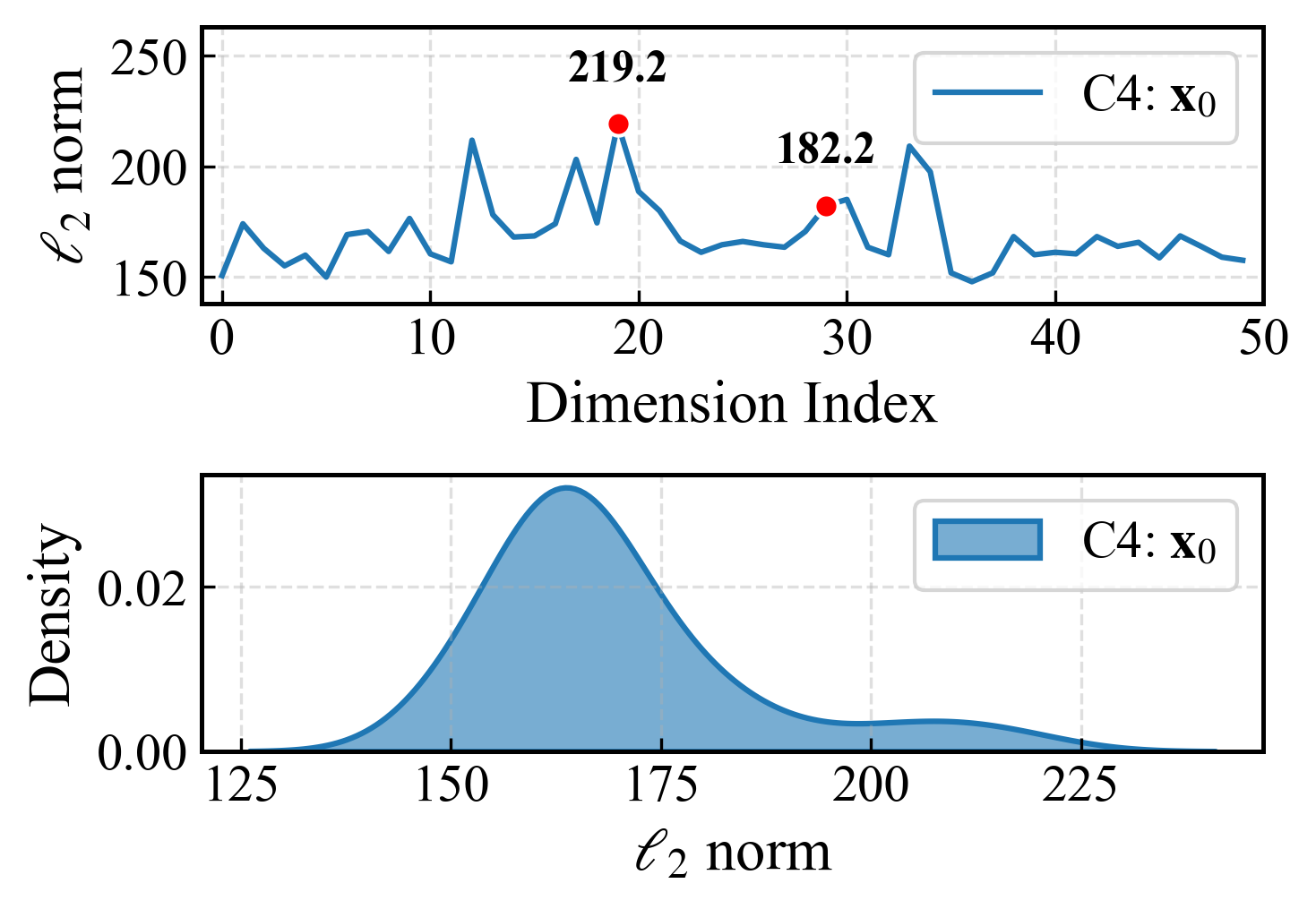}
        \label{subfig:upstream_dist}
    }
\subfloat[Downstream activation distribution]{
        \includegraphics[width=0.325\linewidth]{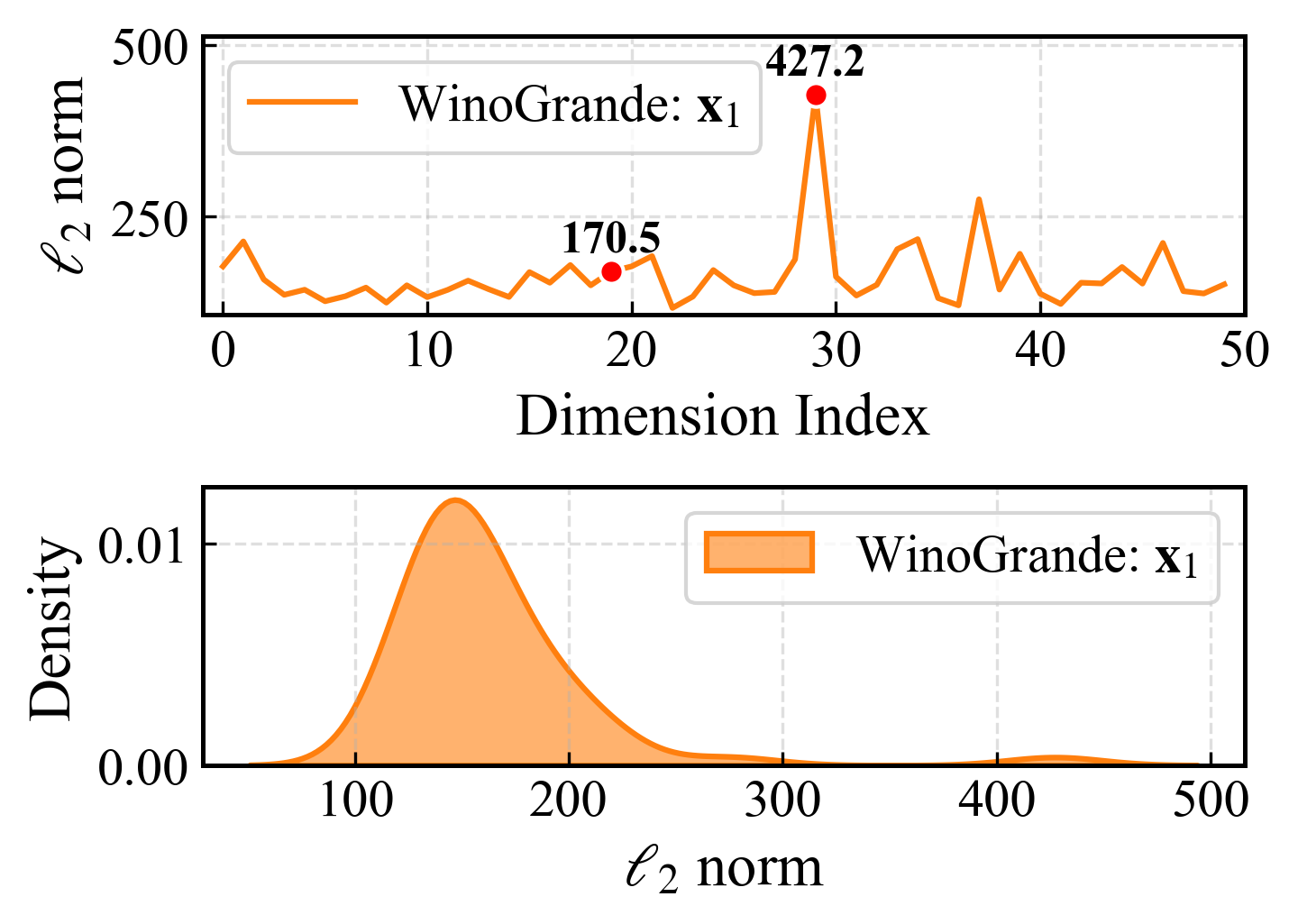}
        \label{subfig:downstream_dist}
    }
\subfloat[Activation shift distribution]{
        \includegraphics[width=0.325\linewidth]{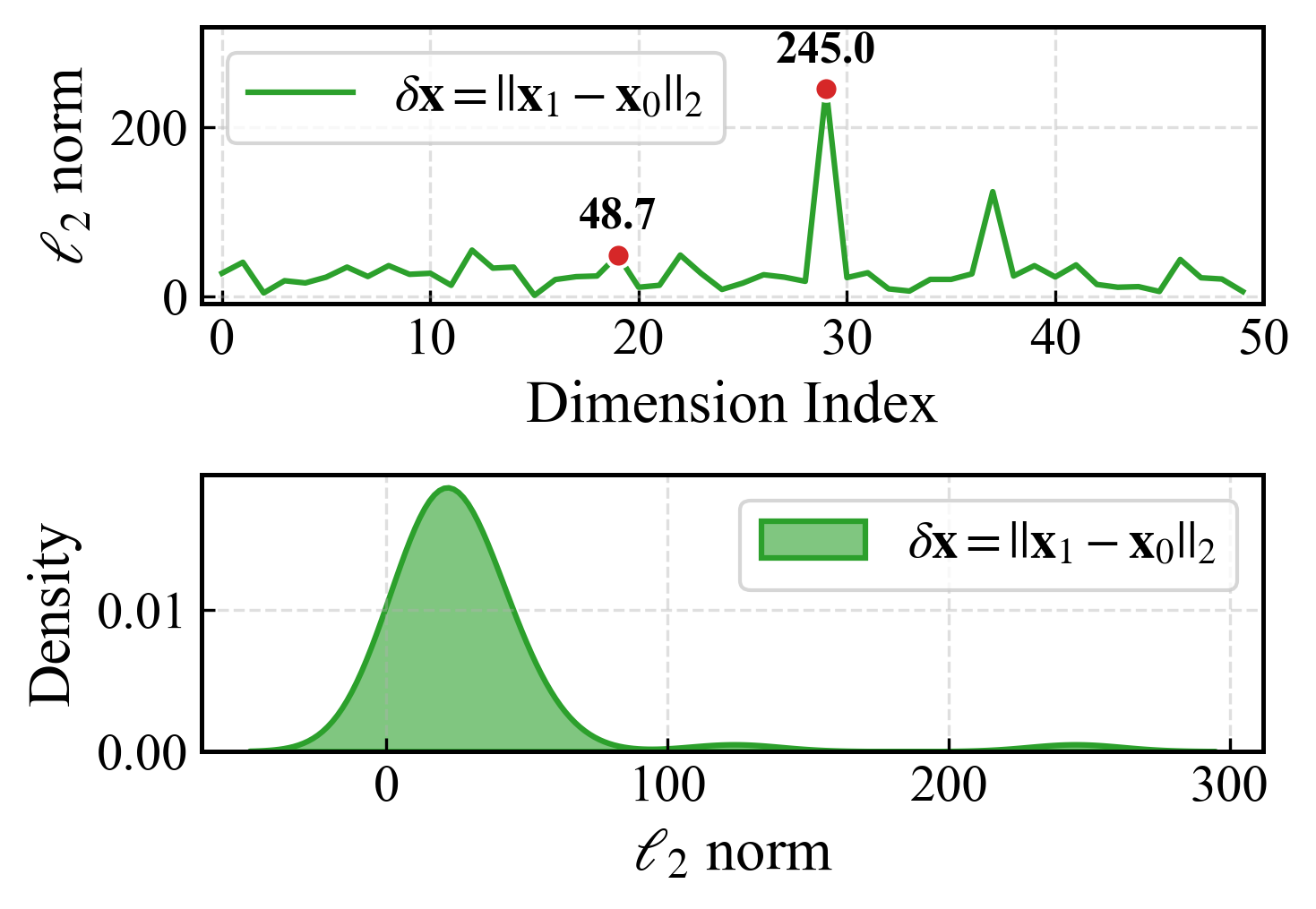}
        \label{subfig:deviation_dist}
    }
\caption{Activation distributions and their shift between upstream and downstream data in LLaMA-2-7B (layer 10). During pruning, a small subset of C4
is used as calibration data (upstream), while downstream tasks such as WinoGrande
are evaluated in a zero-shot setting. 
All inputs are formatted as 128-sample sequences with maximum embedding length. The $\mathrm{L}_2$ norms represent the average activation magnitude across samples.
}
\label{fig:llama-2-7b-activation-shift-measure}
\end{figure*}

Recently, post-training pruning methods~\cite{kwon2022fast, zhang2024plug} have emerged as a more efficient schema for compressing LLMs as they require no additional training resources.
These methods typically design weight importance metrics to remove redundant weights with minimal impact on layer-wise output errors, categorized into non-weight-update pruning (e.g., Magnitude~\cite{han2015learning}, Wanda~\cite{sun2023simple}, Pruner-Zero~\cite{dong2024pruner}) and weight-update pruning (e.g., LLM Surgeon~\cite{van2023llm}, SparseGPT~\cite{frantar2023sparsegpt}, SparseLLM~\cite{bai2024sparsellm}, ADMM-Grad~\cite{bovza2024fast}). Through in-depth analysis, we identify two fundamental limitations in current methodologies stemming from inadequate consideration of activation impacts on error propagation. 

\textbf{First, the constant activation assumption in error estimation.}
Most mainstream pruning methods inherit the “constant activation assumption” from early methods like OBS~\cite{hassibi1993optimal}, which treat input activations as fixed during pruning. This assumption is valid for small-scale networks, where the train/test data is independent and identically distributed. It fails to hold in modern large language models (LLMs). Our experiments reveal significant shifts in input activation across datasets and layers. 
For example, Figure 2 illustrates a segment of activation features from the 10th layer of LLaMA-2-7B. The activation norms exhibit significant shifts, indicating a large deviation between the activations across datasets and layers. 

Non-weight-update methods, such as Wanda and Pruner-Zero, evaluate importance via the product of weight (or gradient) magnitudes and activation norms, which essentially reflects output amplitude rather than true pruning error. Weight-update methods, such as SparseGPT and OBS, explicitly approximate the pruning-induced error via the single-variable-based Taylor expansion on the weight but omit the effect of activation variation between calibration and downstream data, causing inaccurate error modeling. Thus, the fixed activation assumption is fundamentally flawed for LLM pruning. A more accurate modeling of activation shift is required for precise error estimation and robust downstream performance.

\begin{figure*}[!t] 
    \centering
    \includegraphics[width=\linewidth]{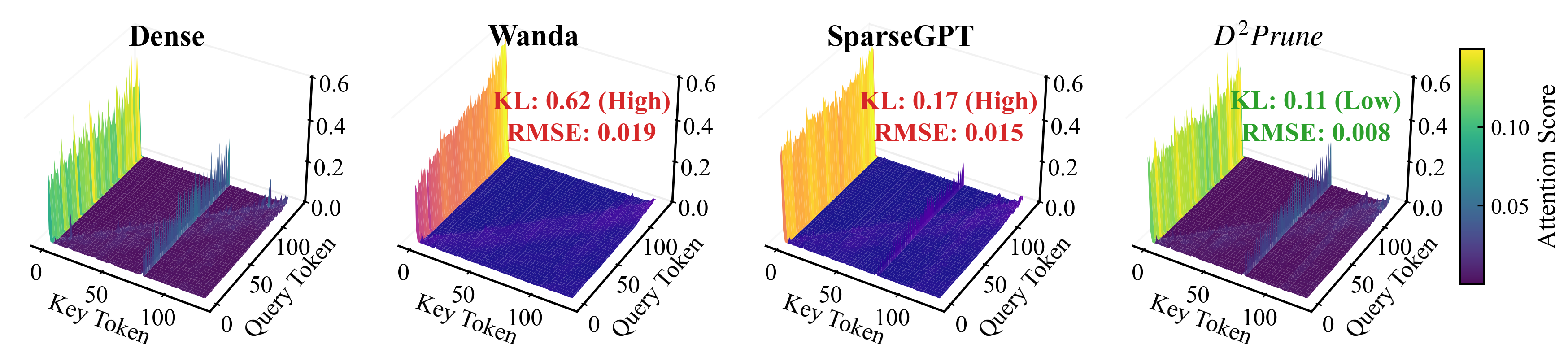} %
    \caption{\textbf{Visualizing of uniformized multi-head attention in LLaMA-2-7B (80\% Sparsity, 128-token sequences from C4 as Calibraion input).} 
    We compare the 3D attention scores of the surfaces in the \textbf{final Transformer layer} for dense models against pruned counterparts (Wanda, SparseGPT, and our $D^2Prune$). \textbf{$D^2Prune$} selectively updates projections to optimize for distribution consistency, successfully preserving these patterns with minimal distortion (lowest KL/RMSE).} 
    \label{fig:attention_surface_llama_2_7b} %
\end{figure*}

\textbf{Second, negligence of long-tail distribution in attention modules.} Multi-head attention, as the core component of Transformers, shows long-tail attention distribution (as noted in several literatures~\cite{zhou2021informer, ji2021distribution, chen2022reltransformer}) where a few key tokens dominate attention scores, which is pivotal for LLM reasoning.  Nonetheless, current pruning methods treat query (Q), key (K), and value (V) weights in attention modules identically to linear layers (Gate, Up, and Down in MLP modules), incurring significant error propagation. Specifically, \textbf{the non-weight-update pruning accumulates errors in Q/K/V layers by retaining raw weights, while weight-update pruning disrupts the long-tail attention distribution through global weight adjustments.} As illustrated in Figure~\ref{fig:attention_surface_llama_2_7b}, direct updates to Q/K/V weights homogenize attention scores from the original long-tail distribution, severely distorting the model’s focus on critical tokens. Conversely, skipping updates leads to significant errors in attention outputs due to error accumulation. 
This "all-or-nothing" approach fails to balance error compensation and distribution preservation, causing catastrophic performance drops, especially at high sparsity. 

To tackle the above limitations, we propose $D^2Prune$, a novel pruning method for large language models via Dual Taylor Expansion and Attention Distribution Awareness. Specifically, to address the first limitation, we propose \textbf{Activation-Weight Dual-Sensitive Pruning Mechanism}. 
By formulating a Dual Taylor Expansion of the error function with respect to both activations and weights, our method jointly models the impacts of activation variations and weight perturbations on output errors, enabling precise pruning mask selection and weight updates.
This dual expansion reduces perplexity by approximately 10\% compared to conventional single-variable approaches 
, and it enhances accuracy at high sparsity by up to 40\% on downstream tasks with significant distribution shifts. To address the second limitation, we propose \textbf{Attention Distribution-Aware Dynamic Weight Update Strategy}. 
Specifically, we formulate Q/K/V update states as a combinatorial optimization problem and propose a perplexity-guided lightweight adaptive search method that dynamically identifies update configurations. This strategy can effectively preserve the attention distribution while reducing the pruning errors by allowing weight updating. 
Experimental results demonstrate that \textbf{this strategy reduces the KL divergence of the attention distribution and the root-mean-square error (RMSE) of attention outputs by 61\% and 43\% on average} compared with non-weight-update (Wanda) and weight-update (SparseGPT) baselines.
Our contributions are summarized as follows: 
\begin{itemize}
    \item We propose a novel pruning framework ($D^2Prune$) for compressing LLMs, which simultaneously accounts for activation dynamics and weight sensitivity via Dual Taylor Expansion, enabling effective pruning mask selection and weight updating. 
    \item We propose an attention distribution-aware dynamic weight update method, which dynamically identifies update configurations for q, k, v weights in attention modules, achieving balance in attention distribution preservation and pruning errors compensation.
    \item  Experimental results demonstrate that $D^2Prune$ achieves SOTA performance across varying parameter scales of LLMs (including OPT-125M, LLaMA2/3 models, Qwen3-8/14B.), with substantial improvements in perplexity and zero-shot accuracy across various sparsity levels, especially at high sparsity. 
\end{itemize}

\section{Preliminaries}
\label{sec:pm}

\subsection{Related Work}
Post-training pruning has emerged as a critical technique for compressing Large Language Models (LLMs) \cite{kwon2022fast, zhang2024plug, reyhan2024novel}. In Appendix G, we review post-training pruning techniques in detail, categorizing them into the non-weight-update and weight-update pruning methods.

\subsection{Problem Statement}
Post-training pruning is performed by decomposing the full-model compression problem into layer-wise pruning subproblems, which can be written as:
\begin{equation}
\label{eq:non-weight-repiar}
\underset{\boldsymbol{M}_l}{\mathrm{arg}\min}||\boldsymbol{W}_l\boldsymbol{X}_l-\left( \boldsymbol{M}_l\odot \boldsymbol{W}_l \right) \boldsymbol{X}_l||_{2}^{2},
\end{equation} 
where the subscript $l$ represents the $l_{th}$ layer, $X_l$ is the activation, $W_l$ is the weight matrix, and $\boldsymbol{M}_l$ is the pruning mask corresponding to the weights, respectively. 
Furthermore, 
weight-update methods aim to adjust the remaining unpruned weights to compensate for the pruning-induced error, resulting in an updated weight matrix $
\hat{\boldsymbol{W}}_l
$.
The optimization objective is as follows:
\begin{equation}
\label{eq:weight-update}
\underset{\boldsymbol{M}_l,\hat{\boldsymbol{W}}_l}{\mathrm{arg}\min}||\boldsymbol{W}_l\boldsymbol{X}_l-\left( \boldsymbol{M}_l\odot \hat{\boldsymbol{W}}_l \right){\boldsymbol{X}_l}||_{2}^{2}. 
\end{equation}

Nonetheless, simultaneously solving for both the sparsity mask and the adjustments to the remaining weights is an NP-hard problem. Mainstream approaches~\cite{frantar2023sparsegpt, bai2024sparsellm, hassibi1993optimal} typically use a pruning metric to select the pruning mask and then optimize the remaining weights based on that mask.
Prior studies mainly adopt the layer-wise pruning error minimization. They usually solve Eq.~\ref{eq:non-weight-repiar} and Eq.~\ref{eq:weight-update} through the single-variable Taylor expansion with respect to weights.
OBS and SparseGPT calculate the inverse of the second-order derivative matrix (Hessian) to perform weight mask selection and update.
Specifically, the output error change of a layer can be written in the single-variable Taylor expansion form in OBS and SparseGPT:
\begin{equation}
\label{eq:single local error}
\delta E=\left( \frac{\partial E}{\partial \mathbf{w}} \right) ^T\delta \mathbf{w}+\frac{1}{2}\delta \mathbf{w}^T\mathbf{H}\delta \mathbf{w},
\end{equation} 
where $\textbf{H}=\partial ^2E/\partial \textbf{w}^2$, ${\delta \textbf{w}}=\textbf{w}-\textbf{w}_0$, ${\delta }E=E\left( \textbf{w} \right) -E\left( \textbf{w}_0 \right)$ represents the change of the objective function (error or loss change).  Following the principle proposed in Optimal Brain Damage~\cite{NIPS1989_6c9882bb}, we define the saliency of a parameter as the increase in the objective function (i.e., loss) after removing that parameter. Therefore, the goal of pruning is to minimize the change in error given by Eq.~\ref{eq:single local error}, when setting one of the weights \textbf{w} to zero (represented as $w_q$).
Elimating $w_q$ is expressed as $\delta {w_q} + {w_q} = 0$ or expressed in vector form more generally, that is $
\textbf{e}_{q}^{T}\textbf{w}+w_q=0
$, where $\textbf{e}_{q}$ is the unit vector in weight space corresponding to scalar $w_q$. 
Then, the goal is to solve:
\begin{equation}
\label{eq:optimal-obs}
\delta \mathbf{w}^* 
= \arg\min_{\delta \mathbf{w}}\;\tfrac{1}{2}\,\delta \mathbf{w}^T \mathbf{H}\,\delta \mathbf{w}
\quad\text{s.t.}\quad
\mathbf{e}_q^T \,\delta \mathbf{w}+w_q=0.
\end{equation}

To solve the Eq.~\ref{eq:optimal-obs}, we can form a Lagrangian function:
\begin{equation}
L_q=\frac{1}{2}\delta \textbf{w}^T\cdot \textbf{H}\cdot \delta \textbf{w}+\lambda \left( \textbf{e}_{q}^{T}\delta \textbf{w}+w_q \right), 
\end{equation}
where $\lambda$ is a Lagrange undetermined multiplier.
Finally, we can obtain the optimal weight change  $\delta \textbf{w}$ and the change in error $\delta E$: 
\begin{equation}
\label{eq:obs results}
{\delta \textbf{w}}=-\frac{w_q}{\left( \boldsymbol{\textbf{H}}^{-1} \right) _{qq}}\left( \boldsymbol{\textbf{H}}^{-1} \right) _{:,q},\delta E=L_q=\frac{w_{q}^{2}}{\left( \boldsymbol{\textbf{H}}^{-1} \right) _{qq}}.
\end{equation} 

Since $L_q$ is the increase in error and is used to reflect the "saliency" of $w_q$ in prior studies, both the weight-update and non-weight-update pruning methods evaluate the importance of $w_q$ utilizing the magnitude $L_q$ or its variants. They rank weights by $L_q$ and then prune the lowest-saliency weights to meet the target sparsity. 

\section{Methodology}
We propose $D^2Prune$, a novel pruning framework for LLMs. An overview of $D^2Prune$ is shown in Figure~\ref{fig:$D^2Prune$ Framework} and Algorithm 1 (available in Appendix A).
\begin{figure*}
\centering
\includegraphics[width=1.0\linewidth]{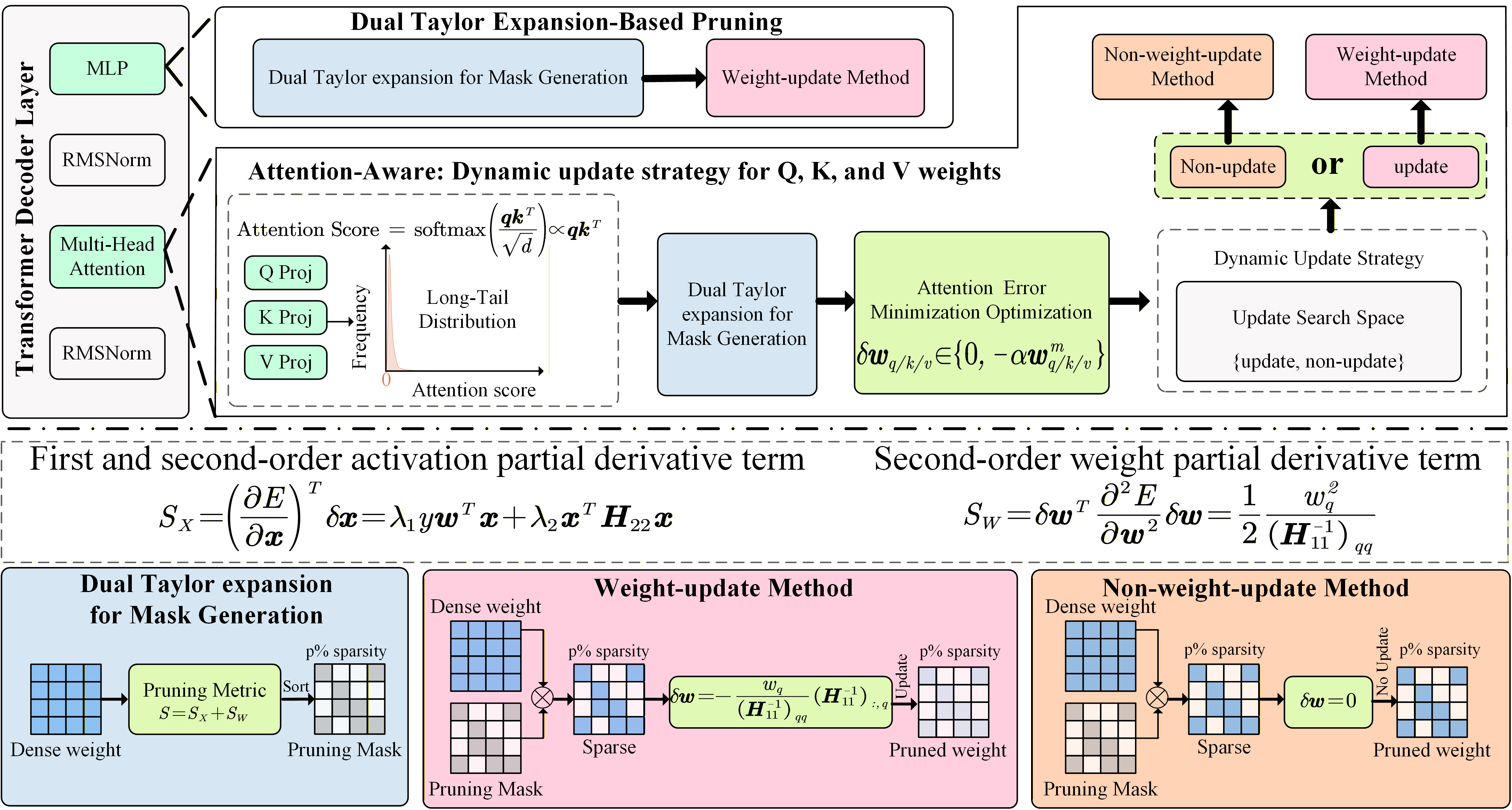}
\caption{Illustration of the $D^2Prune$ framework.
Using the update status of q, k, and v weights in each attention layer as the search space, we take the minimization of ppl as the objective to dynamically adapt the update strategy of the q, k, and v weights.}
\label{fig:$D^2Prune$ Framework}
\end{figure*}

\subsection{Dual Taylor Expansion for Mask Selection \& Weight Update}
\label{sec:Dual Taylor Expansion of local error}
Consider a pre-trained large language model where the original input activation of a certain layer is $\boldsymbol{x}_0$, the weight is $\boldsymbol{w}_0$, and the output is $
y={\boldsymbol{x}_0}^T\boldsymbol{w}_0$. Now, suppose we remove a certain weight such that
the error is $
E=||\boldsymbol{x}^T\boldsymbol{w}-{\boldsymbol{x_0}}^T\boldsymbol{w}_0||_{2}^{2}
$, where $\boldsymbol{w}=\boldsymbol{w}_0+\delta \boldsymbol{w}$ denotes the weights retained after pruning, and $
\boldsymbol{x}$ is the input activation.
\textbf{Previous works typically sample a small calibration set and compute layerwise input activations during pruning, assuming the activations remain constant, i.e., $\boldsymbol{x}=\boldsymbol{x}_0$.}
However, as demonstrated in Figure~\ref{subfig:deviation_dist}, there exists a significant distributional shift between upstream and downstream activations. A more realistic formulation is $\boldsymbol{x}=\boldsymbol{x}_0+\delta \boldsymbol{x}$, where $\delta\boldsymbol{x}$ is activation shift. Neglecting activation shift leads to inaccurate error estimation in prior pruning methods.

To address this issue, we now consider the error minimization problem as a dual function of both weights and activations. 
The first-order partial derivative term with respect to the weights is 0 as a network trained to a local minimum in error, and the third and all higher-order terms converge to 0 and can be eliminated. Therefore, following OBS and SparseGPT (refer to Eq.~\ref{eq:single local error}), the functional Taylor series of the error change with respect to weights and activations is:
\begin{equation}
\delta E=\left( \frac{\partial E}{\partial \boldsymbol{u}} \right) ^T\delta \boldsymbol{u}+\left( \frac{1}{2}\delta \boldsymbol{u}^T\boldsymbol{H}\delta \boldsymbol{u} \right), 
\end{equation}
where $
\boldsymbol{u}=\left( \boldsymbol{w},\boldsymbol{x} \right) ^T
$,$\frac{\partial E}{\partial \boldsymbol{u}}=\left( \frac{\partial E}{\partial \boldsymbol{w}},\frac{\partial E}{\partial \boldsymbol{x}} \right) ^T$,$
\delta \boldsymbol{u}=\left( \delta \boldsymbol{w},\delta \boldsymbol{x} \right) ^T
$, 
$\boldsymbol{H} = \left( \begin{smallmatrix}
\partial^2 E / \partial \boldsymbol{w}^2 & \partial^2 E / \partial \boldsymbol{w} \, \partial \boldsymbol{x} \\
\partial^2 E / \partial \boldsymbol{x} \, \partial \boldsymbol{w} & \partial^2 E / \partial \boldsymbol{x}^2
\end{smallmatrix} \right)$. 
It is worth noting that, due to the presence of Layer Normalization in Transformers, the input activations of each layer are normalized during the forward propagation on the calibration data. Moreover, since the pruning-induced $\delta \boldsymbol{w}$
is small, the cross terms in the Hessian matrix become higher-order infinitesimals and have a negligible impact on error estimation (a theory derivation and experimental verification detailed in Appendix C
). Therefore, it is reasonable to assume that the Hessian is predominantly dominated by its diagonal elements, such that:
\begin{equation}
\label{eq:delta E in x and w}
\delta E = \left( \frac{\partial E}{\partial \boldsymbol{x}} \right)^T \delta \boldsymbol{x}
+ \frac{1}{2} \delta \boldsymbol{w}^T \boldsymbol{H}_{11} \delta \boldsymbol{w}
+ \frac{1}{2} \delta \boldsymbol{x}^T \boldsymbol{H}_{22} \delta \boldsymbol{x},
\end{equation}
where $\frac{\partial E}{\partial \boldsymbol{x}}=-\boldsymbol{w}\left( \boldsymbol{x}^T\boldsymbol{w}-y \right) 
$, 
$
\boldsymbol{H}_{11}=\frac{\partial ^2E}{\partial \boldsymbol{w}^2}=\boldsymbol{xx}^{\boldsymbol{T}}
$, 
$
\boldsymbol{H}_{\boldsymbol{2}\boldsymbol{2}}=\frac{\partial ^2E}{\partial \boldsymbol{x}^2}=\boldsymbol{ww}^{\boldsymbol{T}}
$, 
$
\delta E=E\left( \boldsymbol{u} \right) -E\left( \boldsymbol{u}_0 \right)
$ represents the change of objective function (loss or error). Following the principle proposed in Optimal Brain Damage~\cite{hassibi1993optimal}, it is reasonable to define the saliency of a parameter as the increase in the objective function (i.e., loss) after removing that parameter, since objective functions play a central role in neural network pruning. Therefore, $\delta E$ 
captures the change in the objective function (i.e., network error) caused by perturbations to the weights and activations.

Since the activations of the batch training data are not available for each layer,  computing $
\delta \boldsymbol{x}
$ directly is difficult.  
Fortunately, as shown in Figure 7-9 (detailed in Appendix C, average cosine similarity across layers reaches 0.96, range: 0.94–0.98), we observe a linear correlation between activation norm shifts across different tasks and the calibration activations across model layers (while the relationship may appear nonlinear at the level of individual weight dimensions (see Figure~\ref{fig:llama-2-7b-activation-shift-measure}), the layerwise nature of our error estimation makes the linear approximation reasonably valid. As evidenced by the results in 
Table 12 (detailed in Appendix B.4), 
this assumption leads to improved robustness in practice. We leave more fine-grained modeling of $\delta \boldsymbol{x}$ to future work.

Thus, we assume that there exists a perturbation coefficient $\lambda$ such that $
\boldsymbol{\delta x}=\lambda \boldsymbol{x}
$ (a theoretical analysis with experimental support results detailed in 
Appendix C.
Substituting this into Eq.~\ref{eq:delta E in x and w} and eliminating the constant coefficient term, we can obtain:
\begin{align}
\delta E &= \lambda y\boldsymbol{w}^T\boldsymbol{x} + \left( \frac{1}{2}\lambda^2 - \lambda \right) \boldsymbol{x}^T\boldsymbol{w}\boldsymbol{w}^T\boldsymbol{x} + \frac{1}{2}\delta \boldsymbol{w}^T\boldsymbol{H}_{11} \delta \boldsymbol{w} \notag \\
        &= \lambda_1 y\boldsymbol{w}^T\boldsymbol{x} + \lambda_2 \boldsymbol{x}^T \boldsymbol{H}_{22} \boldsymbol{x} + \frac{1}{2} \delta \boldsymbol{w}^T \boldsymbol{H}_{11} \delta \boldsymbol{w},
\end{align}
where $\lambda _1=\lambda$, $
\lambda _2=\frac{1}{2}\lambda ^2-\lambda 
$ denote the perturbation coefficients of the first-order activation bias and the second-order activation bias term, respectively. 
the goal of pruning  
$w_q$ should satisfy $w_q+\delta w_q=0$, 
i.e., $
\boldsymbol{e}_{q}^{T}\delta \boldsymbol{w}+\boldsymbol{w}=0
$, the solution of $
\delta \boldsymbol{w}
$ and $
\delta E
$ can be obtained using the Lagrange multiplier method as follows: 
\begin{align}
\label{eq:prune-general-formulation}
\delta \boldsymbol{w} &= -\frac{w_q}{\left( \boldsymbol{H}_{11}^{-1} \right)_{qq}} \left( \boldsymbol{H}_{11}^{-1} \right)_{:,q}, \notag \\
L_q &= \lambda_1 y \boldsymbol{w}^T \boldsymbol{x} + \lambda_2 \boldsymbol{x}^T \boldsymbol{H}_{22} \boldsymbol{x} + \frac{1}{2} \frac{w_q^2}{\left( \boldsymbol{H}_{11}^{-1} \right)_{qq}}.
\end{align}

Finally, we use $L_q$ as the pruning metric $S$ to select the pruning mask and update the weights with $\delta \boldsymbol{w}$. It is worth noting that, unlike pruning methods such as SparseGPT, Wanda, and Pruner-Zero, our pruning metric $S$ explicitly incorporates the output activation (i.e., $y$, the input activation of the next layer), demonstrating its ability to model cross-layer dependency.
\subsection{Attention Distribution-Aware Dynamic Weight Update Strategy}
\label{sec:Attention-Aware}
For the q/k/v weights in the attention module, under the pruning mask obtained by dual error minimization and after pruning a certain weight $w_m$, the weights are updated as 
$
{w}_i={w}_{0}^{i}-\delta {w}_i
\left( i\ne m \right)
$,
where $w_{0}^{i}$ denotes the original i-th weight, $
\delta w_i
= \alpha_i\,w_q
(
\alpha_i = (H^{-1})_{qi}/{(H^{-1})_{qq}})
$
. The original attention of the i-th token is:
\begin{equation}
\mathrm{Attention}\left( Q,K,V \right) _i=\sum_{j=1}^n{\mathrm{soft}\max \left( \frac{\boldsymbol{q}_i\boldsymbol{k}_{j}^{T}}{\sqrt{d}} \right)}\boldsymbol{v}_j,
\end{equation}
where $\left( Q,K,V \right)_i$ corresponds to the query, key, and value of the i-th token. $n$ denotes the number of tokens and $d$ is the feature dimension. 
The softmax operator computation result is proportional to the product of $
\boldsymbol{q}_i
$ and the transposed $
\boldsymbol{k}_j
$ and the attention is proportional to the elements of $
\boldsymbol{v}_j
$. 
The long-tail distribution of multi-head attention is critical for semantic understanding in LLMs, where a small fraction of positions carry disproportionately large, critical weights. Unconstrained linear updates to the remaining q/k/v  weights shift these relative magnitudes, distorting the original long-tail pattern and undermining attention fidelity. As illustrated in Figure~\ref{fig:attention_surface_llama_2_7b}, the non-weight-update method can better maintain the long-tail distribution while achieving worse model performance.

However, by entirely refraining from updating pruned q/k/v weights, local errors propagate and accumulate across layers. To address this issue, we propose an attention distribution-aware dynamic weight update strategy that explicitly preserves key attention patterns via a distribution-constrained optimization objective. After pruning off the m-th weight for q, k,v weights of multi-head attention (MHA) and under the pruning mask selection based on the dual error minimization method, 
the restoration of q/k/v weights is equivalent to solving the following optimization problems:
\begin{equation}
\begin{aligned}
\min_{\delta w_{q/k/v}}  
& \mathcal{L}_{q/k/v} + \rho \cdot \mathrm{KL}\left( \mathrm{MHA}(\tilde{w}_{q/k/v}) \mid\mid \mathrm{MHA}(w_{q/k/v}) \right), \\
\text{s.t.} \quad 
& \delta w_{q/k/v} \in \{0, -\alpha_{q/k/v} w_{q/k/v}^{m}\}, \\
& \mathcal{L}_{q/k/v} = \sum_{*\in \{q,k,v\}} \left\| w_* x_* - (M_* \odot \tilde{w}_*) x_* \right\|_2^2, \\
& \tilde{w}_{q/k/v} = w_{q/k/v} + \delta w_{q/k/v},
\end{aligned}
\label{eq: update for qkv}
\end{equation}
where $
L_{q/k/v}
$ measures the output error from pruning for q, k, and v layer. $
{\mathrm{KL}}
$
is the Kullback–Leibler divergence enforcing the pruned attention distribution to mimic the original long-tail pattern. 
$
\rho 
$
is the Lagrangian multiplier, which balances error minimization and distribution preservation.
$\delta \boldsymbol{w}_{q/k/v}$ represents $\sigma \boldsymbol{w}_q$, $\sigma \boldsymbol{w}_k$ or $\sigma \boldsymbol{w}_v$, $\alpha_{q/k/v}$ denotes $\alpha_q$, $\alpha_k$ or $\alpha_v$. The essence of the above optimization problem is to determine whether the q, k, and v weights need to be updated. 

To solve this problem, we explicitly formulate the update strategy of $q$, $k$, and $v$ weights as a constrained optimization problem that balances pruning error minimization and attention distribution preservation. Specifically, we design the search space using the binary update status $\{\text{update}, \text{non-update}\}$ for each of the $q$, $k$, and $v$ weights in attention layers: "update" enables weight adjustment to compensate for pruning-induced errors, while "non-update" preserves the original weight structure to maintain the long-tail attention distribution. To identify the optimal configuration, we adopt perplexity (ppl) as the objective metric, which comprehensively reflects both the model's language modeling performance (linked to pruning error $L_{q/k/v}$) and the fidelity of attention patterns (linked to KL divergence of distributions). This dynamic adaptation process leverages the cross-layer consistency of outlier ratios in $q/k/v$ weights (e.g., 98\% layers in LLaMA-2-13B show the highest outlier ratio in $k$ matrix, see Figure 10 in Appendix D) to reduce the search space, ensuring efficiency while aligning with the dual optimization goal of minimizing error and preserving distribution.

\begin{table*}[!t]
\centering
\resizebox{2\columnwidth}{!}{
\begin{threeparttable}
\begin{small}
\renewcommand{\multirowsetup}{\centering}
\begin{tabular}{cc|cccccccccc}
\toprule
\multicolumn{1}{c}{\multirow{2}{*}{Sparsity}} & \multirow{2}{*}{Method} & \multicolumn{2}{c}{OPT-125M}                               & \multicolumn{2}{c}{LLaMA-2-7B}                           & \multicolumn{2}{c}{LLaMA-2-13B}                          & \multicolumn{2}{c}{LLaMA-2-70B}                  & \multicolumn{2}{c}{LLaMA-3-8B} \\ 
\cmidrule(lr){3-4} \cmidrule(lr){5-6} \cmidrule(lr){7-8} \cmidrule(lr){9-10} \cmidrule(lr){11-12}
\multicolumn{1}{c}{}                          &                         & PPL $\downarrow$              & \multicolumn{1}{c|}{ACC $\uparrow$}                & PPL $\downarrow$            & \multicolumn{1}{c|}{ACC $\uparrow$}                                     & PPL $\downarrow$            & \multicolumn{1}{c|}{ACC $\uparrow$}                                     & PPL $\downarrow$    & \multicolumn{1}{c|}{ACC $\uparrow$}                                     & PPL $\downarrow$       & ACC $\uparrow$                \\ 
\toprule
\multicolumn{1}{c|}{0}                        & Dense                   & 27.66            & \multicolumn{1}{c|}{39.68}              & 5.12           & \multicolumn{1}{c|}{64.38}              & 4.57           & \multicolumn{1}{c|}{67.06}              & 3.12   & \multicolumn{1}{c|}{71.50}              & 5.54      & 68.40              \\ 
\midrule
\multicolumn{1}{c|}{\multirow{4}{*}{50}}                       & SparseGPT               & 36.85            & \multicolumn{1}{c|}{39.82}              & 6.52           & \multicolumn{1}{c|}{60.35}              & 5.63           & \multicolumn{1}{c|}{64.87}              & 3.98   & \multicolumn{1}{c|}{71.35}              & 8.56      & 63.13              \\
\multicolumn{1}{c|}{}                         & Wanda                   & 38.88            & \multicolumn{1}{c|}{39.75}              & 6.44           & \multicolumn{1}{c|}{60.46}              & 5.58           & \multicolumn{1}{c|}{64.17}              & 3.98   & \multicolumn{1}{c|}{70.96}              & 9.06      & 61.02              \\
\multicolumn{1}{c|}{}                         & Pruner-Zero             & 38.80            & \multicolumn{1}{c|}{39.09}              & 6.43           & \multicolumn{1}{c|}{59.26}              & 5.57           & \multicolumn{1}{c|}{64.21}              & -      & \multicolumn{1}{c|}{-}                  & 8.52      & 60.28              \\
\multicolumn{1}{c|}{}                         & \textbf{$D^2Prune$ (Ours)}                 & \textbf{34.98}   & \multicolumn{1}{c|}{\textbf{40.39}}     & \textbf{6.36}  & \multicolumn{1}{c|}{\textbf{61.06}}     & \textbf{5.53}           & \multicolumn{1}{c|}{\textbf{65.90}}              & \textbf{3.93}   & \multicolumn{1}{c|}{\textbf{71.60}}              & \textbf{8.34}      & \textbf{63.58}              \\ 
\midrule
\multicolumn{1}{c|}{\multirow{4}{*}{60}}      & SparseGPT               & 59.46            & \multicolumn{1}{c|}{39.37}              & 9.56            & \multicolumn{1}{c|}{55.34}              & 7.77           & \multicolumn{1}{c|}{60.26}              & 4.98   & \multicolumn{1}{c|}{70.09}              & 14.40     & 55.39              \\
\multicolumn{1}{c|}{}                         & Wanda                   & 74.39            & \multicolumn{1}{c|}{39.45}              & 9.89           & \multicolumn{1}{c|}{53.91}              & 7.87           & \multicolumn{1}{c|}{59.57}              & 4.99   & \multicolumn{1}{c|}{69.04}              & 22.80     & 48.12              \\
\multicolumn{1}{c|}{}                         & Pruner-Zero             & 68.07            & \multicolumn{1}{c|}{39.32}              & 10.34          & \multicolumn{1}{c|}{52.52}              & 7.82           & \multicolumn{1}{c|}{58.12}              & -      & \multicolumn{1}{c|}{-}                  & 20.30     & 50.65              \\
\multicolumn{1}{c|}{}                         & \textbf{$D^2Prune$ (Ours)}                 & \textbf{52.10}   & \multicolumn{1}{c|}{\textbf{40.14}}     & \textbf{9.05}  & \multicolumn{1}{c|}{\textbf{56.08}}     & \textbf{7.49}           & \multicolumn{1}{c|}{\textbf{60.81}}              & 4.88   & \multicolumn{1}{c|}{\textbf{70.65}}              & \textbf{13.44}     & \textbf{56.95}              \\ 
\midrule
\multicolumn{1}{c|}{\multirow{4}{*}{70}}      & SparseGPT               & 218.29           & \multicolumn{1}{c|}{36.31}              & 29.62          & \multicolumn{1}{c|}{44.75}              & 18.20          & \multicolumn{1}{c|}{48.12}              & 8.61   & \multicolumn{1}{c|}{63.49}              & 38.85     & 43.56              \\
\multicolumn{1}{c|}{}                         & Wanda                   & 347.42           & \multicolumn{1}{c|}{35.43}              & 84.92          & \multicolumn{1}{c|}{36.68}              & 44.86          & \multicolumn{1}{c|}{39.75}              & 40.27  & \multicolumn{1}{c|}{60.44}              & 114.30    & 37.07              \\
\multicolumn{1}{c|}{}                         & Pruner-Zero             & 317.87           & \multicolumn{1}{c|}{36.88}              & 151.92         & \multicolumn{1}{c|}{38.09}              & 44.76          & \multicolumn{1}{c|}{43.33}              & -      & \multicolumn{1}{c|}{-}                  & 280.33    & 36.17              \\
\multicolumn{1}{c|}{}                         & \textbf{$D^2Prune$ (Ours)}                 & \textbf{160.81}  & \multicolumn{1}{c|}{\textbf{38.09}}     & \textbf{21.10} & \multicolumn{1}{c|}{\textbf{47.97}}     & \textbf{16.51} & \multicolumn{1}{c|}{\textbf{48.32}}              & \textbf{8.17}   & \multicolumn{1}{c|}{\textbf{64.06}}              & \textbf{33.37}     & \textbf{44.82}              \\ 
\midrule
\multicolumn{1}{c|}{\multirow{4}{*}{80}}      & SparseGPT               & 2140.55          & \multicolumn{1}{c|}{35.08}              & 102.43         & \multicolumn{1}{c|}{36.23}              & 99.14          & \multicolumn{1}{c|}{38.28}              & 25.86  & \multicolumn{1}{c|}{47.71}              & 178.01    & 36.95              \\
\multicolumn{1}{c|}{}                         & Wanda                   & 1920.63          & \multicolumn{1}{c|}{34.93}              & 5107.20        & \multicolumn{1}{c|}{33.72}              & 1384.40        & \multicolumn{1}{c|}{34.67}              & 156.68 & \multicolumn{1}{c|}{37.98}              & 2245.91   & 34.87              \\
\multicolumn{1}{c|}{}                         & Pruner-Zero             & 1251.38          & \multicolumn{1}{c|}{35.06}              & 10244.70       & \multicolumn{1}{c|}{34.83}              & 2040.65        & \multicolumn{1}{c|}{35.09}              & -      & \multicolumn{1}{c|}{-}                  & 10420.01  & 35.31              \\
\multicolumn{1}{c|}{}                         & \textbf{$D^2Prune$ (Ours)}                 & \textbf{1038.87} & \multicolumn{1}{c|}{\textbf{36.29}}     & \textbf{92.68} & \multicolumn{1}{c|}{\textbf{39.09}}     & \textbf{76.80} & \multicolumn{1}{c|}{\textbf{39.42}}              & \textbf{21.37}  & \multicolumn{1}{c|}{\textbf{48.09}}              & \textbf{151.47}    & \textbf{38.73}              \\ 
\midrule
\multicolumn{1}{c|}{\multirow{4}{*}{2:4}}     & SparseGPT               & 59.76            & \multicolumn{1}{c|}{\multirow{4}{*}{-}} & 10.18          & \multicolumn{1}{c|}{\multirow{4}{*}{-}} & 8.39           & \multicolumn{1}{c|}{\multirow{4}{*}{-}} & 5.32   & \multicolumn{1}{c|}{\multirow{4}{*}{-}} & 14.16     & \multirow{4}{*}{-} \\
\multicolumn{1}{c|}{}                         & Wanda                   & 79.80            & \multicolumn{1}{c|}{}                   & 11.35          & \multicolumn{1}{c|}{}                   & 8.37           & \multicolumn{1}{c|}{}                   & 5.18   & \multicolumn{1}{c|}{}                   & 22.86     &                    \\
\multicolumn{1}{c|}{}                         & Pruner-Zero             & 70.92            & \multicolumn{1}{c|}{}                   & 11.16          & \multicolumn{1}{c|}{}                   & 8.05           & \multicolumn{1}{c|}{}                   & -      & \multicolumn{1}{c|}{}                   & 23.56     &                    \\
\multicolumn{1}{c|}{}                         & \textbf{$D^2Prune$ (Ours)}                 & \textbf{59.43}            & \multicolumn{1}{c|}{}                   & \textbf{10.00}          & \multicolumn{1}{c|}{}                   & \textbf{8.02}           & \multicolumn{1}{c|}{}                   & \textbf{5.12}   & \multicolumn{1}{c|}{}                   & \textbf{14.10}     &                    \\ 
\midrule
\multicolumn{1}{c|}{\multirow{4}{*}{3:4}}     & SparseGPT               & 1365.58          & \multicolumn{1}{c|}{\multirow{4}{*}{-}} & 154.23         & \multicolumn{1}{c|}{\multirow{4}{*}{-}} & 147.23         & \multicolumn{1}{c|}{\multirow{4}{*}{-}} & 54.84  & \multicolumn{1}{c|}{\multirow{4}{*}{-}} & 281.74    & \multirow{4}{*}{-} \\
\multicolumn{1}{c|}{}                         & Wanda                   & 2497.68          & \multicolumn{1}{c|}{}                   & 3111.14        & \multicolumn{1}{c|}{}                   & 5815.71        & \multicolumn{1}{c|}{}                   & 386.57 & \multicolumn{1}{c|}{}                   & 13054.17  &                    \\
\multicolumn{1}{c|}{}                         & Pruner-Zero             & 2946.15          & \multicolumn{1}{c|}{}                   & 7913.18        & \multicolumn{1}{c|}{}                   & 4134.56        & \multicolumn{1}{c|}{}                   & -      & \multicolumn{1}{c|}{}                   & 854085.30 &                    \\
\multicolumn{1}{c|}{}                         & \textbf{$D^2Prune$ (Ours)}                 & \textbf{1346.36}          & \multicolumn{1}{c|}{}                   & \textbf{136.89}         & \multicolumn{1}{c|}{}                   & \textbf{143.69}         & \multicolumn{1}{c|}{}                   & \textbf{48.06}  & \multicolumn{1}{c|}{}                   & \textbf{190.35}    &                    \\ 
\bottomrule
\end{tabular}
\end{small}
\end{threeparttable}
}
\caption{Pruning comparison across language modeling and zero-shot tasks at different sparsity levels (middle: 50\%, 60\% and 2:4, high: 70\%, 80\% and 3:4). 
}
\label{tab:language modeling and zero-shot}
\end{table*}
\section{Experiments}\label{sec:experiments}
\subsection{Experimental Settings}
We compare $D^2Prune$ with three main baselines, including weight-update pruning methods such as SparseGPT~\cite{frantar2023sparsegpt} and non-weight-update pruning methods such as Wanda~\cite{sun2023simple} and Pruner-Zero~\cite{dong2024pruner}, and all experiments are conducted on an NVIDIA A40 GPU. Moreover, given the potential of our dynamic attention update mechanism for global modeling, we also compare our method with a global pruning approach, SparseLLM~\cite{bai2024sparsellm}, which iteratively updates activations and weights in the MLP modules to minimize global error detailed results in the Appendix B.5). In addition, we further evaluate our method on the latest large language models Qwen3-8B and 14B~\cite{qwen}, as well as the vision transformer model DeiT~\cite{touvron2021training} built on the ViT~\cite{dosovitskiy2020image} architecture detailed in Appendix B.3). Following prior studies, we assess and compare various pruning methods based on language modeling with the perplexity metric (ppl) and the zero-shot accuracy for downstream tasks.
For all pruning algorithms, we use the C4~\cite{raffel2020exploring} training set as the calibration dataset, and we use 128 samples and segment the pre-trained LLMs according to their maximum embedding dimensions. For the language modeling comparison, we assess the perplexity on the WikiText2~\cite{merity2016pointer} test set.
For the zero-shot comparison, we use seven benchmark tasks  including BoolQ~\cite{clark2019boolq}, HellaSwag~\cite{zellers2019hellaswag}, WinoGrande, RTE~\cite{wang2018glue}, ARC-c and ARC-e (ARC Easy and Challeng~\cite{clark2018think}), and OBQA~\cite{mihaylov2018can} from the EleutherAI LM Harness~\cite{gao2023framework}. All the datasets are sourced from the HuggingFace Datasets library. 

\subsection{Experimental Results}
\subsubsection{Unified Evaluation on Language Modeling and Zero-Shot.}
Following SparseGPT, Wanda and Pruner-Zero, we provide a unified pruning performance evaluation result in Table~\ref{tab:language modeling and zero-shot} on both language modeling (in terms of perplexity, PPL↓) and zero-shot tasks (measured by average accuracy, ACC↑) under various sparsity levels and model sizes. 
Overall, $D^2Prune$ consistently outperforms existing pruning baselines across both moderate (50\%–60\%, 2:4) and high sparsity (70\%–80\%, 3:4) regimes. Compared to weight-update methods (e.g., SparseGPT), $D^2Prune$ achieves on average a 3.1\% accuracy gain and 16\% lower perplexity. Against non-weight-update methods (e.g., Wanda, Pruner-Zero), the perplexity reduction reaches up to 86\%.
Notably, $D^2Prune$ surpasses the dense model on LLaMA-2-70B at 50\% sparsity in zero-shot accuracy, demonstrating strong generalization. Due to resource limitations, Pruner-Zero fails to perform pruning on LLaMA-2-70B even with a single A40 or A100 GPU. Therefore, the corresponding entries are marked as “–” in the table. For semi-structured pruning (e.g., 2:4 and 3:4 patterns), we report only the perplexity results, while accuracy is marked as “–” to indicate that it was not evaluated. The detailed zero-shot results on seven downstream tasks are shown in Appendix B.1, providing a comprehensive understanding of the zero-shot capabilities for the discussed models. 

\subsubsection{Expanding Experiment Results.}
(1) We further compare with the more computationally intensive \textbf{global pruning} method SparseLLM~\cite{bai2024sparsellm} (detailed in Appendix B.5 and (2) validate $D^2Prune$ on \textbf{vision transformers} (e.g., DeiT~\cite{touvron2021training}) and \textbf{recent LLMs like Qwen3~\cite{qwen}} (detailed in Appendix B.3, confirming its broad applicability across architectures and tasks. (3) In addition, to comprehensively evaluate the reasoning capabilities of pruned models in real-world scenarios, we conduct \textbf{in-context learning (ICL) experiments on the GSM8K} benchmark~\cite{cobbe2021training} (detailed in Appendix B.2). The results demonstrate that in the few-shot setting, $D^2Prune$ consistently outperforms baselines. (3) Finally, we provide \textbf{pruning efficiency and speedup} experiment results in the Appendix E. 

\subsection{Ablation study}
\subsubsection{Effectiveness of the Dual Taylor Expansion for mask selection 
\& weight update.}
To validate the effectiveness of the dual Taylor expansion for mask selection and weight update, 
%
we incorporated the first and second order activation partial derivatives to SparseGPT and Wanda for mask selection and weight update, and we denote these methods with the prefix ``$D^2$-''. 
As shown in Table~\ref{tab:du-sparsegpt-wanda}, $D^2$-SparseGPT consistently outperforms SparseGPT across all models and sparsity levels, and $D^2$-Wanda consistently outperforms Wanda, demonstrating the effectiveness of the dual Taylor expansion. 
Besides, we perform the ablation study on key hyperparameters $\lambda_1$,  $\lambda_2$ and $s$ for dual Taylor expansion in 
Appendix B.4. 
\begin{table}[t]
\centering
\scriptsize
\setlength{\tabcolsep}{3.0pt}
\begin{threeparttable}
\begin{small}
\renewcommand{\multirowsetup}{\centering}
\begin{tabular}{c|cccc}
\toprule
Method & 50 & 60 & 70 & 80 \\ \midrule
SparseGPT & 8.56 & 14.40 & 38.85 & 178.01 \\
$D^2$-Sparsegpt & \textbf{8.48} & \textbf{13.66} & \textbf{38.64} & \textbf{170.92} \\ \midrule
Wanda & 9.06 & 22.80 & 114.30 & 2245.91 \\
$D^2$-Wanda & \textbf{8.92} & \textbf{22.54} & \textbf{106.58} & \textbf{1414.32} \\ 
\bottomrule
\end{tabular}
\end{small}
\end{threeparttable}
\caption{Ablations study of dual Taylor expansions on LLaMA-3-8B (WikiText2 perplexity, unstructured 50\%-80\% sparsity, Dense: 5.54). 
``$D^2$-'' indicates the first- and second-order activation partial derivative term 
$S_X$ (see 
Algorithm 1, detailed in Appendix A
).
}
\label{tab:du-sparsegpt-wanda}
\end{table}

\subsubsection{Effectiveness of Attention Distribution-Aware Dynamic Weight
Update Strategy.}
As shown in Table~\ref{tab:hybrid pruning}, we perform unstructured pruning on the LLaMA-2-7B. 
Specifically, when either the q, k, or v weight is not updated (denoted as w/o q, w/o k, and w/o v), the pruning performance outperforms Wanda and is competitive with SparseGPT.
Furthermore, our attention distribution-aware dynamic weight update strategy dynamically sets one of the q, k, and v weights to the non-update state while keeping the others in the update state, leading to optimal performance for all LLMs.
This strongly supports the effectiveness of the attention distribution-aware dynamic weight update strategy. 
\begin{table}[t]
\centering
\scriptsize              
\setlength{\tabcolsep}{1.5pt}  
\begin{threeparttable}
\begin{small}
\begin{tabular}{cc|cccc}
\toprule
Method & Weight-Update Layers & 50 & 60 & 70 & 80 \\ \midrule
\multicolumn{1}{c|}{SparseGPT} & ALL Layers & 6.52 & 9.56 & 29.62 & 102.43 \\
\multicolumn{1}{c|}{Wanda} & None & 6.44 & 9.89 & 84.92 & 5107.20 \\
\multicolumn{1}{c|}{Pruner-Zero} & None & 6.43 & 10.34 & 151.92 & 10244.70 \\ \midrule
\multicolumn{1}{c|}{\multirow{6}{*}{$D^2Prune$}} & ALL Layers & 6.46 & 9.46 & 23.45 & 101.88 \\
\multicolumn{1}{c|}{} & None & 6.43 & 9.75 & 81.62 & 3561.51 \\ 
\multicolumn{1}{c|}{} & w/o q & 6.50 & 9.60 & 25.16 & 104.63 \\
\multicolumn{1}{c|}{} & w/o k & 6.49 & 9.55 & 24.04 & 118.79 \\
\multicolumn{1}{c|}{} & w/o v & 6.36 & 9.05 & 21.10 & 92.68 \\
\multicolumn{1}{c|}{} & Dynamic Update & \textbf{6.36} & \textbf{9.05} & \textbf{21.10} & \textbf{92.68} \\ 
\bottomrule
\end{tabular}
\end{small}
\end{threeparttable}
\caption{Ablations study of dynamic weight update for attention modules (q, k, v) on LLaMA-2-7B (WikiText2 perplexity, unstructured 50\%-80\% sparsity, Dense: 5.12). }
\label{tab:hybrid pruning}
\end{table}

\subsection{Analysis}
\subsubsection{Impact of sparsity on the update configurations of q, k, v weights.}
When searching for optimal q, k, and v update configurations, we evaluated various settings under different sparsity levels. 
The configuration yielding the lowest perplexity was selected as the optimal setup. The results in Table \ref{tab:hybrid pruning} and 
Appendix D 
reveal a key insight: the optimal q, k, and v update configuration remains consistent across all sparsity levels for each model, indicating that it is independent of sparsity. 

\subsubsection{LoRA Fine-tuning.}
We evaluated the impact of fine-tuning on the pruned models.
Specifically, we employ LoRA~\cite{hu2022lora} (r=8, $\alpha$=16) for fine-tuning with one epoch on the C4 training datasets (training sample size is 30k) using one GPU (40G memory) and 15 hours. We conducted experiments on LLaMA-2-7B, involving unstructured pruning at 60\% and 70\% sparsity levels. 
Table \ref{tab:llama-2-7b-lora-ft} reports the perplexity on the WikiText2 dataset and mean accuracy of zero-shot tasks. 
We can observe that LoRA fine-tuning can effectively restore the performance of all pruned models, and $D^2Prune$ consistently outperforms all other pruning methods in both accuracy and perplexity.
This further demonstrates the effectiveness of $D^2Prune$.
\begin{table}[t]
\centering
\scriptsize
\setlength{\tabcolsep}{3.5pt}
\begin{threeparttable}
\begin{small}
\begin{tabular}{cc|cc|cc}
\toprule
\multirow{2}{*}{Method} & \multirow{2}{*}{Fine-tuning} & \multicolumn{2}{c|}{60} & \multicolumn{2}{c}{70} \\ 
\cmidrule{3-4} \cmidrule{5-6}
 &  & PPL & ACC & PPL & ACC \\ \midrule
\multicolumn{1}{c|}{\multirow{2}{*}{SparseGPT}} & \xmark & 9.56 & 55.34 & 29.62 & 44.75 \\
\multicolumn{1}{c|}{} & LoRA & 7.18 & 57.82 & 10.18 & 49.73 \\ \midrule
\multicolumn{1}{c|}{\multirow{2}{*}{Wanda}} & \xmark & 9.86 & 53.91 & 84.92 & 36.68 \\
\multicolumn{1}{c|}{} & LoRA & 7.26 & 56.45 & 11.81 & 46.43 \\ \midrule
\multicolumn{1}{c|}{\multirow{2}{*}{$D^2Prune$}} & \xmark & \textbf{9.05} & \textbf{56.06} & \textbf{21.10} & \textbf{47.97} \\
\multicolumn{1}{c|}{} & LoRA & \textbf{7.05} & \textbf{60.59} & \textbf{9.70} & \textbf{51.37} \\ \bottomrule
\end{tabular}
\end{small}
\end{threeparttable}
\caption{Fine-tuning can mitigate the perplexity gap to dense LLM. $D^2Prune$ significantly outperforms all other pruning approaches even after LoRA fine-tuning.}
\label{tab:llama-2-7b-lora-ft}
\end{table}

\section{Conclusion}
We propose \textbf{$D^2Prune$}, a novel pruning algorithm for compressing large language models (LLMs).
By extending error minimization with dual Taylor expansion and attention distribution awareness, $D^2Prune$ improves the accuracy of pruning mask selection and weight updates.
It explicitly models the dual effects of activation and weight variations, and introduces a dynamic weight update strategy for Transformer attention modules to reduce attention distribution errors.
Extensive experiments show that $D^2Prune$ consistently outperforms recent post-training pruning methods.
\section{Acknowledgments}
This work is supported by the National Natural Science Foundation of China (Nos. 62572085 and 62472058).


\bibliography{main}

\appendix
\setcounter{secnumdepth}{2}
\onecolumn
\begin{algorithm*}[th!]

\caption{$D^2Prune$: Dynamic Dual-layer Pruning with Selective Q/K/V Update. We prune all the layer weight matrix of FFN or MLP by weight-update method (SparseGPT~\cite{frantar2023sparsegpt} solver based on Dual Taylor Expansion), and prune the layer weight matrix $W_q$, $W_k$ or $W_v$ of multi-head attention (MHA) by non-weight-update method (Wanda~\cite{sun2023simple} solver based on Dual Taylor Expansion)
detailed in Appendix~\ref{q,k,v update seach}).  
We apply the scaling factor $s$ to both input and output activations in the activation derivative terms to standardize and balance the weight magnitudes (detailed in Appendix~\ref{sec:Hyperparameter Selection for $D^2Prune$}). For example, 
$
||a||_2\gets s\left( ||a||_2 \right) =\left( ||a||_{2}^{2}/s \right) ^{1/2}
$ for activations $a$. 
}
\label{alg:$D^2Prune$}
\resizebox{1\columnwidth}{!}{
\begin{threeparttable}
\begin{small}
\renewcommand{\multirowsetup}{\centering}
\SetKwInput{KwInput}{Input}
\SetKwInput{KwOutput}{Output}
\DontPrintSemicolon

\KwInput{Transformer layer with MHA and FFN/MLP modules. For FFN/MLP: weight matrix $FC1$/$W_{\text{up}}, $FC2$/W_{\text{down}}$. For MHA: weight matrix $W_q$, $W_k$, $W_v$, $W_o$, input activations $X$, output activations $Y$, target sparsity $p\%$, hyperparameters $\lambda_1$, $\lambda_2$, scaling factor $s$.}

\SetKwFunction{PruneFFN}{\textbf{$D^2Prune$ on FFN}}
\SetKwProg{Fn}{Function}{:}{}
\Fn{\PruneFFN{$W_{\text{up}}, W_{\text{down}}, X, Y$}}{
    Apply scaling: $\|X\|_2 \gets \left( \|X\|_2^2 / s \right)^{1/2}$, $\|Y\|_2 \gets \left( \|Y\|_2^2 / s \right)^{1/2}$ \\
    \ForEach{linear layer $l \in \{\text{up}, \text{down}\}$}{
     \tcp{Prune by weight-update method $\gets$ SparseGPT Solver based on Dual Taylor expansion}
        Compute mask $M_l$ selecting $(1 - p\%)$ weights $w_q \in W_l$ with highest scores: \\
        \Indp $\lambda_1 \|Y\|_2 \cdot |W| \cdot \|X\|_2 + \lambda_2 |W|^2 \cdot \|X\|_2^2 + \frac{1}{2}\frac{w_q^2}{(H_{11}^{-1})_{qq}}$ \\
        
        Pruning weights: $W_l  \gets (1-M_l) \cdot W_l$ \\ 
        
        Update weights: $W_l \gets W_l - \frac{w_q}{(H_{11}^{-1})_{qq}} \cdot (H_{11}^{-1})_{:,q}$ \\
    }
    \KwRet $W_{\text{up}}, W_{\text{down}}$
}

\SetKwFunction{PruneMHA}{\textbf{$D^2Prune$ on MHA}}
\SetKwProg{Fn}{Function}{:}{}
\Fn{\PruneMHA{$W_q, W_k, W_v, W_o, X, Y$}}{
    \ForEach{$l_{\text{non}} \in \{q, k, v\}$}{
        \ForEach{$l \in \{q, k, v, o\}$}{
            \uIf{$l == l_{\text{non}}$}{
                \tcp{Prune by non-weight-update method $\gets$ Wanda Solver based on Dual Taylor expansion}
                Compute mask $M_l$ selecting $(1 - p\%)$ weights $w_q \in W_l$ with highest scores: \\
                \Indp $\lambda_1 \|Y\|_2 \cdot |W| \cdot \|X\|_2 + \lambda_2 |W|^2 \cdot \|X\|_2^2 + |W|^2 \cdot \|X\|_2^2$
                \\
                Pruning weights: $W_l  \gets (1-M_l) \cdot W_l$
            }
            \Else{
                \tcp{ Prune by weight-update method $\gets$ SparseGPT Solver based on Dual Taylor expansion}
                Compute mask $M_l$ selecting $(1 - p\%)$ weights $w_q \in W_l$ with highest scores:  \\
                \Indp $\lambda_1 \|Y\|_2 \cdot |W| \cdot \|X\|_2 + \lambda_2 |W|^2 \cdot \|X\|_2^2 + \frac{1}{2}\frac{w_q^2}{(H_{11}^{-1})_{qq}}$ \\
                
                Pruning weights: $W_l  \gets (1-M_l) \cdot W_l$ \\
                
                Update weights: $W_l \gets W_l - \frac{w_q}{(H_{11}^{-1})_{qq}} \cdot (H_{11}^{-1})_{:,q}$
            }
        }
        Apply pruned $\{W_q, W_k, W_v\}$ to MHA module and evaluate model performance (e.g., PPL). \\
    }
    Select $\argmin$-PPL combination of $\{q, k, v\}$ as $l_{\text{non}}^*$. \\
    \KwRet Optimal $\{W_q, W_k, W_v\}$ under $l_{\text{non}}^*$, $W_o$
}
\end{small}
\end{threeparttable}
}
\end{algorithm*}
\twocolumn
\section{$D^2Prune$ Algorithm Pseudo-Code}
\label{sec:Pseudo-Code}
In Section 3 (detailed in main text), we provide the detailed theoretical derivation of $D^2Prune$. We can unify the weight importance function $S$ for pruning mask selection into the following formula:
\begin{equation}
\vspace{-1mm}
    S=f\left( X,Y,W \right) =S_X+S_W,
\end{equation}
where $S_X$ is the first and second-order partial derivative term of the error with respect to activations, and $S_W$ is the second-order partial derivative term of the error with respect to weights, satisfying:
\begin{equation}
S_X=\lambda _1YWX+\lambda _2X^TH_{22}X,S_W=\frac{1}{2}\frac{W^2}{H_{11}^{-1}}.
\end{equation}
Note that $H_{22} = WW^T$ and $H_{11} = XX^T$.
To avoid the complex matrix inversion required for computing the Hessian and to accelerate the calculation of $D^2\text{Prune}$, we simplify $S_W$ in the non-weight-update pruning method by following the idea proposed in Wanda\cite{sun2023simple}, resulting in $S_W = \frac{1}{2}W^2 X^2$.
Additionally, inspired by quantization methods such as SmoothQuant \cite{xiao2023smoothquant} and OmniQuant \cite{shao2023omniquant}, we introduce scaling factors to balance the magnitudes of weights and activations. This scaling operation is crucial to appropriately reflect the importance of both components during pruning without compromising pruning accuracy. Such that:
\begin{equation}
\begin{aligned}
S_X &= \lambda_1 s_x^1(Y) \cdot s_x^2(W) \cdot s_x^1(X)
      + \lambda_2 s_x^2(W) \cdot s_x^1(X), \\
S_W &= s_w^2(W) \cdot s_w^1(X).
\end{aligned}
\end{equation}
where $s_x=\left( s_{x}^{1},s_{x}^{2} \right) $
and $s_w=\left( s_{w}^{1},s_{w}^{2} \right) $ are the scaling factors for the first-order activation partial derivative term and the second-order weight partial derivative term, respectively. Specifically, $s_{a}^{1}$ is regularized using the L2 norm and scaled by a certain magnitude $s$ ($s=1500$ in this paper for all the experiments) to prevent excessive dominance due to scaling differences, while $s_{a}^{2}$ is regularized using the L1 norm to measure the relative magnitude of the weights. In Algorithm~\ref{alg:$D^2Prune$}, we provide the detailed pseudocode for $D^2Prune$.
\begin{table*}[htbp]
\centering
\resizebox{2.1\columnwidth}{!}{
\begin{threeparttable}
\renewcommand{\multirowsetup}{\centering}
\begin{tabular}{cc|cccccccccccccccccccc}
\toprule
                                                  &                          & \multicolumn{4}{c}{OPT-125M}                                                                                                                       & \multicolumn{4}{c}{LLaMA-2-7B}                                                                                 & \multicolumn{4}{c}{LLaMA-2-13B}                                                        & \multicolumn{4}{c}{LLaMA-2-70B}                                                        & \multicolumn{4}{c}{LLaMA-3-8B}                                             \\ 
                                                  \cmidrule(lr){3-6} \cmidrule(lr){7-10} \cmidrule(lr){11-14} \cmidrule(lr){15-18} \cmidrule(lr){19-22}
\multirow{-2}{*}{Dataset}                         & \multirow{-2}{*}{Method} & 50                            & 60                            & 70                            & \multicolumn{1}{c|}{80}                            & 50                                     & 60             & 70             & \multicolumn{1}{c|}{80}             & 50             & 60             & 70             & \multicolumn{1}{c|}{80}             & 50             & 60             & 70             & \multicolumn{1}{c|}{80}             & 50             & 60             & 70             & {80} \\ 
\toprule
\multicolumn{1}{c|}{}                             & SparseGPT                & 61.19 & 60.42 & 41.22 & \multicolumn{1}{c|}{39.32} & 76.52          & \textbf{73.42} & 65.08          & \multicolumn{1}{c|}{45.93}          & 82.05          & 78.77          & \textit{66.81} & \multicolumn{1}{c|}{60.91}          & \textbf{84.52} & 84.22          & 80.67          & \multicolumn{1}{c|}{69.91}          & 77.06          & 75.4           & \textbf{67.52} & 47.98                   \\
\multicolumn{1}{c|}{}                             & Wanda                    & \cellcolor[HTML]{EFEFEF}62.11 & 61.46 & 39.02 & \multicolumn{1}{c|}{37.88} & 76.51          & 66.39          & 42.29          & \multicolumn{1}{c|}{37.83}          & 81.68          & 77.15          & 62.32          & \multicolumn{1}{c|}{37.83}          & 83.12          & 83.67          & 74.25          & \multicolumn{1}{c|}{62.20}          & \textbf{79.32} & 66.05          & 49.36          & 37.98                   \\
\multicolumn{1}{c|}{}                             & Pruner-Zero              & 61.37 & 59.35 & 44.98 & \multicolumn{1}{c|}{38.01} & 72.84          & 65.22          & 45.30          & \multicolumn{1}{c|}{37.83}          & 80.94          & 77.12          & 62.72          & \multicolumn{1}{c|}{37.83}          & -              & -              & -              & \multicolumn{1}{c|}{-}              & 76.36          & 68.74          & 38.13          & 37.83                   \\
\multicolumn{1}{c|}{\multirow{-4}{*}{BoolQ}}      & \textbf{$D^2Prune$ (Ours)}                  & \textbf{62.17}                & \textbf{61.59}                & \textbf{50.40}                & \multicolumn{1}{c|}{\textbf{40.89}}                & \textbf{76.61} & 73.03          & \textbf{66.79} & \multicolumn{1}{c|}{\textbf{61.16}} & \textbf{81.90} & \textbf{78.78} & 66.79          & \multicolumn{1}{c|}{\textbf{62.02}} & 84.46          & \textbf{84.86} & \textbf{80.83} & \multicolumn{1}{c|}{\textbf{72.54}} & 79.30          & \textbf{77.16} & 67.22          & \textbf{56.18}          \\ 
\midrule
\multicolumn{1}{c|}{}                             & SparseGPT                & 29.77                         & 29.03                         & \textbf{28.47}                & \multicolumn{1}{c|}{26.46}                         & 70.51                                  & 61.84          & 40.16          & \multicolumn{1}{c|}{29.60}          & 75.31          & \textbf{67.57} & \textbf{46.52} & \multicolumn{1}{c|}{29.54}          & 81.52          & \textbf{78.93} & 68.46          & \multicolumn{1}{c|}{\textbf{43.65}} & 71.59          & 59.63          & \textbf{37.65} & 27.78                   \\
\multicolumn{1}{c|}{}                             & Wanda                    & 30.28                         & 29.21                         & 27.87                         & \multicolumn{1}{c|}{26.21}                         & 70.70                                  & 58.51          & 29.93          & \multicolumn{1}{c|}{26.63}          & \textbf{76.11} & 66.04          & 30.70          & \multicolumn{1}{c|}{27.10}          & 81.32          & 77.80          & 63.80          & \multicolumn{1}{c|}{29.06}          & 67.82          & 46.50          & 29.33          & 27.53                   \\
\multicolumn{1}{c|}{}                             & Pruner-Zero              & \textbf{30.46}                & \textbf{29.72}                & 27.93                         & \multicolumn{1}{c|}{26.04}                         & 69.52                                  & 56.90          & 29.74          & \multicolumn{1}{c|}{26.18}          & 74.68          & 65.44          & 36.27          & \multicolumn{1}{c|}{26.27}          & -              & -              & -              & \multicolumn{1}{c|}{-}              & 68.21          & 51.55          & 29.98          & 27.53                   \\
\multicolumn{1}{c|}{\multirow{-4}{*}{HellaSwag}}  & \textbf{$D^2Prune$ (Ours)}                  & 30.10                         & 29.42                         & 28.26                         & \multicolumn{1}{c|}{\textbf{26.90}}                & \textbf{70.88}                         & \textbf{62.05} & \textbf{45.81} & \multicolumn{1}{c|}{\textbf{29.81}} & 76.01          & 67.48          & 46.23          & \multicolumn{1}{c|}{\textbf{29.81}} & \textbf{81.54} & 78.85          & \textbf{68.69} & \multicolumn{1}{c|}{43.18}          & \textbf{72.13} & \textbf{60.06} & 38.48          & \textbf{28.25}          \\ 
\midrule
\multicolumn{1}{c|}{}                             & SparseGPT                & 52.88                         & 52.40                         & \textbf{52.95}                & \multicolumn{1}{c|}{50.59}                         & 69.85                                  & 65.66          & 57.14          & \multicolumn{1}{c|}{48.61}          & 71.43          & 69.85          & 60.61          & \multicolumn{1}{c|}{48.38}          & 78.05          & \textbf{78.69} & 75.69          & \multicolumn{1}{c|}{\textbf{61.88}} & 72.21          & \textbf{67.87} & \textbf{55.25} & 49.72                   \\
\multicolumn{1}{c|}{}                             & Wanda                    & 52.57                         & 52.24                         & 51.38                         & \multicolumn{1}{c|}{49.72}                         & 68.19                                  & 65.11          & 52.33          & \multicolumn{1}{c|}{\textbf{50.67}} & 70.87          & 65.50          & 50.04          & \multicolumn{1}{c|}{50.27}          & 77.51          & 76.16          & 73.80          & \multicolumn{1}{c|}{48.22}          & 69.77          & 59.03          & 48.54          & 50.04                   \\
\multicolumn{1}{c|}{}                             & Pruner-Zero              & 50.35                         & 51.69                         & 51.85                         & \multicolumn{1}{c|}{49.32}                         & 67.96                                  & 62.83          & 51.14          & \multicolumn{1}{c|}{50.12}          & 70.79          & 65.67          & 54.14          & \multicolumn{1}{c|}{\textbf{50.59}} & -              & -              & -              & \multicolumn{1}{c|}{-}              & 69.37          & 60.14          & 50.43          & 50.67                   \\
\multicolumn{1}{c|}{\multirow{-4}{*}{WinoGrande}} & \textbf{$D^2Prune$ (Ours)}                  & \textbf{52.89}                & \textbf{52.96}                & 52.01                         & \multicolumn{1}{c|}{\textbf{53.12}}                & \textbf{70.17}                         & \textbf{66.54} & \textbf{60.30} & \multicolumn{1}{c|}{50.30}          & \textbf{72.14} & \textbf{70.80} & \textbf{61.17} & \multicolumn{1}{c|}{50.36}          & \textbf{78.37} & 78.14          & \textbf{75.69} & \multicolumn{1}{c|}{61.72}          & \textbf{72.61} & 67.48          & 54.93          & \textbf{49.80}          \\ 
\midrule
\multicolumn{1}{c|}{}                             & SparseGPT                & 49.10                         & 49.09                         & 50.90                         & \multicolumn{1}{c|}{52.70}                         & 53.43                                  & 53.43          & 54.15          & \multicolumn{1}{c|}{53.07}          & 63.89          & 56.32          & 53.06          & \multicolumn{1}{c|}{52.70}          & 70.75          & 71.84          & 62.45          & \multicolumn{1}{c|}{57.04}          & 62.81          & 56.31          & 52.70          & 52.70                   \\
\multicolumn{1}{c|}{}                             & Wanda                    & 48.01                         & 49.45                         & 52.34                         & \multicolumn{1}{c|}{51.62}                         & 53.79                                  & 53.42          & 52.70          & \multicolumn{1}{c|}{45.85}          & 58.12          & 59.56          & 52.70          & \multicolumn{1}{c|}{52.70}          & 72.92          & 69.31          & 61.01          & \multicolumn{1}{c|}{49.46}          & 59.20          & 52.70          & 52.70          & 52.70                   \\
\multicolumn{1}{c|}{}                             & Pruner-Zero              & 44.76                         & 48.73                         & 52.70                         & \multicolumn{1}{c|}{\textbf{53.43}}                & 55.60                                  & 53.43          & 53.07          & \multicolumn{1}{c|}{52.70}          & 62.81          & 57.40          & 52.70          & \multicolumn{1}{c|}{52.70}          & -              & -              & -              & \multicolumn{1}{c|}{-}              & 55.96          & 52.71          & 52.70          & 52.70                   \\
\multicolumn{1}{c|}{\multirow{-4}{*}{RTE}}        & \textbf{$D^2Prune$ (Ours)}                  & \textbf{49.46}                & \textbf{50.54}                & \textbf{53.43}                & \multicolumn{1}{c|}{53.07}                         & \textbf{55.60}                         & \textbf{54.51} & \textbf{54.87} & \multicolumn{1}{c|}{\textbf{53.43}} & \textbf{66.43} & \textbf{61.37} & \textbf{53.43} & \multicolumn{1}{c|}{\textbf{52.71}} & \textbf{72.20} & \textbf{74.00} & \textbf{64.62} & \multicolumn{1}{c|}{\textbf{57.40}} & \textbf{64.26} & \textbf{61.37} & \textbf{54.87} & \textbf{52.71}          \\ 
\midrule
\multicolumn{1}{c|}{}                             & SparseGPT                & 23.12                         & 23.21                         & 21.67                         & \multicolumn{1}{c|}{21.93}                         & \textbf{42.49}                         & 34.72          & 25.94          & \multicolumn{1}{c|}{22.35}          & 45.13          & \textbf{42.66} & \textbf{27.64} & \multicolumn{1}{c|}{22.86}          & 55.80          & 53.15          & 44.80          & \multicolumn{1}{c|}{\textbf{27.47}} & \textbf{45.13} & 34.72          & 23.80          & 22.18                   \\
\multicolumn{1}{c|}{}                             & Wanda                    & 22.44                         & 21.58                         & 21.33                         & \multicolumn{1}{c|}{23.46}                         & 42.40                                  & 33.53          & 22.27          & \multicolumn{1}{c|}{24.14}          & 46.07          & 40.35          & 20.95          & \multicolumn{1}{c|}{24.48}          & 55.55          & 52.13          & 41.21          & \multicolumn{1}{c|}{21.93}          & 43.51          & 29.18          & 22.01          & 22.61                   \\
\multicolumn{1}{c|}{}                             & Pruner-Zero              & 22.78                         & 22.69                         & 21.92                         & \multicolumn{1}{c|}{22.44}                         & 39.51                                  & 33.45          & 22.95          & \multicolumn{1}{c|}{\textbf{26.02}} & 44.11          & 37.54          & 24.57          & \multicolumn{1}{c|}{25.34}          & -              & -              & -              & \multicolumn{1}{c|}{-}              & 44.88          & 32.59          & 21.08          & 23.55                   \\
\multicolumn{1}{c|}{\multirow{-4}{*}{ARC-c}}      & \textbf{$D^2Prune$ (Ours)}                  & \textbf{24.15}                & \textbf{23.63}                & \textbf{21.93}                & \multicolumn{1}{c|}{\textbf{23.81}}                & 42.24                                  & \textbf{35.58} & \textbf{27.30} & \multicolumn{1}{c|}{25.17}          & \textbf{46.84} & 41.13          & 27.56          & \multicolumn{1}{c|}{\textbf{25.17}} & \textbf{56.31} & \textbf{54.01} & \textbf{45.48} & \multicolumn{1}{c|}{26.95}          & 44.20          & \textbf{36.34} & \textbf{24.40} & \textbf{24.15}          \\ 
\midrule
\multicolumn{1}{c|}{}                             & SparseGPT                & 37.03                         & \textbf{36.83}                & 32.95                         & \multicolumn{1}{c|}{29.12}                         & 68.69                                  & 61.11          & 42.21          & \multicolumn{1}{c|}{28.40}          & 71.04          & \textbf{65.03} & 49.20          & \multicolumn{1}{c|}{28.53}          & \textbf{80.76} & 78.58          & 70.75          & \multicolumn{1}{c|}{44.44}          & \textbf{71.50} & 57.36          & 40.19          & \textbf{30.47}          \\
\multicolumn{1}{c|}{}                             & Wanda                    & 36.67                         & 35.18                         & 32.28                         & \multicolumn{1}{c|}{28.40}                         & 69.06                                  & 61.40          & 30.81          & \multicolumn{1}{c|}{27.52}          & 71.54          & 64.77          & 32.70          & \multicolumn{1}{c|}{26.68}          & 78.91          & 77.57          & 70.41          & \multicolumn{1}{c|}{28.20}          & 67.71          & 52.57          & 30.72          & 27.61                   \\
\multicolumn{1}{c|}{}                             & Pruner-Zero              & 37.07                         & 36.82                         & \textbf{33.75}                & \multicolumn{1}{c|}{\textbf{29.33}}                & 68.39                                  & 59.81          & 36.45          & \multicolumn{1}{c|}{26.34}          & 71.33          & 65.23          & 44.90          & \multicolumn{1}{c|}{27.10}          & -              & -              & -              & \multicolumn{1}{c|}{-}              & 68.77          & 55.81          & 35.48          & 28.49                   \\
\multicolumn{1}{c|}{\multirow{-4}{*}{ARC-e}}      & \textbf{$D^2Prune$ (Ours)}                  & \textbf{37.58}                & 36.41                         & 33.42                         & \multicolumn{1}{c|}{29.25}                         & \textbf{69.11}                         & \textbf{61.62} & \textbf{47.94} & \multicolumn{1}{c|}{\textbf{28.49}} & \textbf{72.35} & 64.90          & \textbf{49.28} & \multicolumn{1}{c|}{\textbf{29.29}} & 80.72          & \textbf{79.29} & \textbf{71.34} & \multicolumn{1}{c|}{\textbf{44.44}} & 69.78          & \textbf{59.43} & \textbf{43.06} & 29.84                   \\ \hline
\multicolumn{1}{c|}{}                             & SparseGPT                & 25.60                         & 24.60                         & 26.00                         & \multicolumn{1}{c|}{25.40}                         & 41.00                                  & 37.20          & 28.60          & \multicolumn{1}{c|}{25.60}          & 45.20          & 65.03          & 49.20          & \multicolumn{1}{c|}{28.53}          & 48.00          & 45.20          & 41.60          & \multicolumn{1}{c|}{29.60}          & 41.60          & 36.40          & 27.80          & 27.80                   \\
\multicolumn{1}{c|}{}                             & Wanda                    & 26.20                         & 27.00                         & 23.80                         & \multicolumn{1}{c|}{\textbf{27.20}}                & 42.60                                  & 39.00          & 26.40          & \multicolumn{1}{c|}{23.40}          & 44.80          & 64.77          & 32.70          & \multicolumn{1}{c|}{26.68}          & 47.40          & 46.60          & 38.60          & \multicolumn{1}{c|}{26.80}          & 39.80          & 30.80          & 26.80          & 25.60                   \\
\multicolumn{1}{c|}{}                             & Pruner-Zero              & \textbf{26.80}                & 26.20                         & 25.00                         & \multicolumn{1}{c|}{26.80}                         & 41.00                                  & 36.00          & 28.00          & \multicolumn{1}{c|}{\textbf{26.18}} & 44.80          & 65.23          & 44.90          & \multicolumn{1}{c|}{27.10}          & -              & -              & -              & \multicolumn{1}{c|}{-}              & 38.40          & 33.00          & 25.40          & 26.40                   \\
\multicolumn{1}{c|}{\multirow{-4}{*}{OBQA}}       & \textbf{$D^2Prune$ (Ours)}                  & 26.40                         & \textbf{26.40}                & \textbf{27.20}                & \multicolumn{1}{c|}{27.00}                         & \textbf{42.80}                         & \textbf{39.20} & \textbf{32.80} & \multicolumn{1}{c|}{25.20}          & \textbf{45.60}          & \textbf{64.90} & \textbf{49.28} & \multicolumn{1}{c|}{\textbf{29.29}} & \textbf{47.60} & \textbf{45.40} & \textbf{41.80} & \multicolumn{1}{c|}{\textbf{30.40}} & \textbf{42.80} & \textbf{36.80} & \textbf{30.80} & \textbf{30.20}          \\ 
\midrule
\multicolumn{1}{c|}{}                             & SparseGPT                & 39.82                         & 39.37                         & 36.31                         & \multicolumn{1}{c|}{35.08}                         & 60.35                                  & 55.34          & 44.75          & \multicolumn{1}{c|}{36.23}          & 64.87          & 60.26 & 48.12          & \multicolumn{1}{c|}{38.28}          & 71.35          & 70.09          & 63.49          & \multicolumn{1}{c|}{47.71}          & 63.13          & 55.39          & 43.56          & 36.95                   \\
\multicolumn{1}{c|}{}                             & Wanda                    & 39.75                         & 39.45                         & 35.43                         & \multicolumn{1}{c|}{34.93}                         & 60.46                                  & 53.91          & 36.68          & \multicolumn{1}{c|}{33.72}          & 64.17          & 59.57          & 39.75          & \multicolumn{1}{c|}{34.67}          & 70.96          & 69.04          & 60.44          & \multicolumn{1}{c|}{37.98}          & 61.02          & 48.12          & 37.07          & 34.87                   \\
\multicolumn{1}{c|}{}                             & Pruner-Zero              & 39.09                         & 39.32                         & 36.88                         & \multicolumn{1}{c|}{35.06}                         & 59.26                                  & 52.52          & 38.09          & \multicolumn{1}{c|}{34.83}          & 64.21          & 58.12          & 43.33          & \multicolumn{1}{c|}{35.09}          & -              & -              & -              & \multicolumn{1}{c|}{-}              & 60.28          & 50.65          & 36.17          & 35.31                   \\
\multicolumn{1}{c|}{\multirow{-4}{*}{Mean}}       & \textbf{$D^2Prune$ (Ours)}                  & \textbf{40.39}                & \textbf{40.14}                & \textbf{38.09}                & \multicolumn{1}{c|}{\textbf{36.29}}                & \textbf{61.06}                         & \textbf{56.08} & \textbf{47.97} & \multicolumn{1}{c|}{\textbf{39.09}} & \textbf{65.90} & \textbf{60.81}          & \textbf{48.32} & \multicolumn{1}{c|}{\textbf{39.42}} & \textbf{71.60} & \textbf{70.65} & \textbf{64.04} & \multicolumn{1}{c|}{\textbf{48.09}} & \textbf{63.58} & \textbf{56.95} & \textbf{44.82} & \textbf{38.73}          \\ 
\bottomrule
\end{tabular}
\end{threeparttable}
}
\caption{Zero-shot accuracy for each task and their mean accuracy (\%) under varying sparsity (50, 60, 70, 80 (\%)) for OPT-125M and LLaMA-2/3 Models.}
\label{all zero-shot task acc}
\end{table*}

\section{Expanding the Experimental Results}
\subsection{Expanding the Zero-Shot Taks}
\label{sec:Expanding the Zero-Shot Taks}
\textbf{Detailed Results}. Table~\ref{all zero-shot task acc} reports the detailed zero-shot results for OPT-125M, LLaMA-2-7B, LLaMA-2-13B, LLaMA-2-70B, and LLaMA-3-8B across 7 downstream tasks including BoolQ~\cite{clark2019boolq}, HellaSwag~\cite{zellers2019hellaswag}, WinoGrande, RTE~\cite{wang2018glue}, ARC-c and ArC-e (ARC Easy and Challeng~\cite{clark2018think}), and OBQA~\cite{mihaylov2018can}, respectively, providing a comprehensive understanding of the zero-shot capabilities for the discussed models. For a fair comparison, all methods are kept with the same setup, and the evaluation tools are consistent, with the same lm\_eval version (v0.4.4). Since the pruning results for LLaMA-2-70B with Pruner-Zero could not be reproduced on a single GPU (NVIDIA A100). 

\textbf{Analysis and Discussions}. Overall, for the zero-shot accuracy, it can be seen that although there are certain differences across tasks, $D^2Prune$ significantly outperforms non-weight-update pruning methods such as Wanda and Pruner-Zero in most tasks under medium and high sparsity conditions. Compared to weight-update methods, under the medium sparsity condition, the performance of $D^2Prune$ is close to SparseGPT. However, as sparsity increases, $D^2Prune$ outperforms SparseGPT by a large margin, with performance improvements becoming more noticeable and obvious. This may be due to the significant improvement of $D^2Prune$ in the accuracy of local error estimation at high sparsity compared to other pruning methods. Furthermore, it is worth noting that at certain sparsity levels, the pruned model’s performance on downstream tasks can even exceed the Dense baseline. This indicates that pruning involves removing redundant or less important parameters from the model, which can act as a form of regularization or denoising. Similar to dropout, pruning can regularize and eliminate noise in the Dense model, thereby improving generalization capability. This provides strong evidence for the potential of model pruning that offers an effective solution for LLMs compression. For example, As is shown in Table 1 (detailed in main text), under 50\% and 60\% sparsity conditions of OPT-125M, all pruning algorithms outperform the Dense baseline on the BoolQ dataset, with $D^2Prune$ achieving the best results, surpassing Dense by 12\% and 11\%, respectively. Additionally, on the RTE dataset, $D^2Prune$ outperforms Dense under both 50\% and 60\% sparsity, while other pruning algorithms performed slightly worse than Dense, further proving the superiority of our proposed method. Finally, we can also observe that SparseGPT and $D^2Prune$ outperform Dense under 50\% and 60\% sparsity condition for LLaMA-2-70B pruning.

\subsection{Evaluation of In-Context Learning}
To comprehensively evaluate the reasoning capabilities of pruned models in real-world scenarios, we conduct in-context learning (ICL) experiments on the GSM8K benchmark. GSM8K\cite{cobbe2021training} consists of grade-school math word problems that require strong logical reasoning and multi-step arithmetic, making it a challenging testbed for language models. 
Following prior studies~\cite{sun2023simple, dong2024pruner}, we perform 5-shot unstructured ICL evaluations on LLaMA-2-7B and LLaMA-2-13B. In Table~\ref{tab:gsm8k}, we report the accuracy of the dense LLMs and their pruned counterparts under 60\% and 70\% sparsity levels.
The results demonstrate that in the few-shot setting, $D^2Prune$ consistently outperforms existing pruning methods, including SparseGPT~\cite{frantar2023sparsegpt}, Wanda~\cite{sun2023simple}, and Pruner-Zero~\cite{dong2024pruner}, highlighting its superior ability to preserve reasoning performance under aggressive compression.

\begin{table}[htbp]
\centering
\centering
\begin{threeparttable}
\begin{small}
\renewcommand{\multirowsetup}{\centering}
\begin{tabular}{ccc}
\toprule
Dataset     & \multicolumn{2}{c}{GSM8K}     \\ \cline{2-3} 
Model       & LLaMA-2-7B    & LLaMA-2-13B   \\ \hline
Dense       & 13.95         & 23.28         \\ \hline
SparseGPT   & 2.58          & 4.17          \\
Wanda       & 2.20          & 3.79          \\
Pruner-Zero & 1.82          & 3.03          \\
$D^2Prune$     & \textbf{5.84} & \textbf{4.47} \\ \bottomrule
\end{tabular}
\end{small}
\end{threeparttable}
\vspace{-3mm}
\caption{In-Context Learning Accuracy on the GSM8K Dataset}
\label{tab:gsm8k}
\end{table}

\subsection{Additional LLMs and Vision Model Pruning}
In this appendix, we explore whether $D^2Prune$ generalizes to other language and vision models. We apply it to the latest Qwen3~\cite{qwen} series of large language models (8B and 14B), as well as the widely used DeiT~\cite{touvron2021training} model based on the ViT architecture. We report perplexity on WikiText2 for Qwen3-8B/14B and top-1 accuracy on ImageNet-1K for DeiT. As shown in Table~\ref{tab:qwen and deit}, $D^2Prune$ consistently outperforms other pruning methods, further demonstrating its effectiveness and generality.

\begin{table}[t]
\centering
\resizebox{1\columnwidth}{!}{
\begin{threeparttable}
\begin{small}
\renewcommand{\multirowsetup}{\centering}
\setlength{\tabcolsep}{1.45pt}
\begin{tabular}{ccccc}
\hline
\multicolumn{2}{c}{Evaluation}  & \multicolumn{2}{c}{WikiText: Perplextity} & ImageNet: Accuracy \\ \cline{3-5} 
\multicolumn{2}{c}{Model}       & Qwen3-8B            & Qwen3-14B           & DeiT               \\ \hline
\multicolumn{2}{c}{Dense}       & 8.56                & 7.50                & 87.03              \\ \hline
\multirow{3}{*}{60} & SparseGPT & 12.72               & 10.42               & 86.24              \\
                    & Wanda     & 14.49               & 11.11               & 83.32              \\
                    & $D2Prune$   & \textbf{12.31}      & \textbf{10.10}      & \textbf{87.14}     \\ \hline
\multirow{3}{*}{80} & SparseGPT & 75.20               & 66.00               & 72.37              \\
                    & Wanda     & 409.34              & 339.80              & 28.71              \\
                    & $D2Prune$   & \textbf{72.54}      & \textbf{48.15}      & \textbf{74.76}     \\ \bottomrule
\end{tabular}
\end{small}
\end{threeparttable}
}
\caption{Evaluation of $D2Prune$ on Language (Qwen3) and Vision (DeiT) Models}
\label{tab:qwen and deit}
\end{table}

\subsection{Hyperparameter Selection for $D^2Prune$}
\label{sec:Hyperparameter Selection for $D^2Prune$}
\textbf{Perturbation coefficients: $\lambda_1$ and $\lambda_2$}. First, During the pruning mask selection based on the Dual Taylor Expansion of error, there are hyperparameters $\lambda_1$ and $\lambda_2$ (in Eq. (10) (detailed in main text) that represent the perturbation coefficients of the first-order activation bias and the second-order activation bias term, respectively. In the range of values, we set the value of $\lambda_1$ and $\lambda_2$ from 0 to 1 (It is worth noting that in our implementation, a negative sign is applied to $\lambda_2$. Therefore, we focus on the relative variation of the coefficients rather than their signs). Specifically, the parameter search space is set to [0, 0.2, 0.5, 0.8, 1], and we perform a study within this range of hyperparameters to explore their impact on pruning performance. Results for OPT-125M, LLaMA-2-7B, LLaMA-2-13B and LLaMA-3-8B with 80\% sparsity are shown in Table~\ref{tab:models-r1-r2}. We can see that none of the hyperparameter experiments achieved the lowest perplexity at ($\lambda_1$,$\lambda_2$)=(0,0), which further emphasizes the effectiveness of the activation partial derivatives in pruning mask selection. This key result also confirms our hypothesis: since activations are computed from randomly selected mini-batches of calibration data, they inevitably differ from training data in both task type and data distribution. Such discrepancies have a non-negligible impact on layer-wise error estimation during pruning. It is important to note that the theoretical origin of these perturbation coefficients is determined by the differences in activation distributions between the training and calibration datasets, as well as between the calibration and downstream task data. Therefore, by introducing tunable hyperparameters, we aim to guide practitioners in adapting the pruning model based on task distribution discrepancies, which has practical significance. 

For instance, when the downstream task closely matches the calibration dataset, we recommend the default setting of $\lambda = 1$ (i.e., $\lambda_1 = 1$, $\lambda_2 = 0$), where the second-order activation perturbation is small and can be treated as a higher-order infinitesimal. In contrast, when there is a substantial mismatch—such as using language modeling data (e.g., C4 or WikiText2) for calibration but evaluating on more complex tasks involving contextual or commonsense reasoning (e.g., HellaSwag)—adjusting $\lambda_1$ and $\lambda_2$ can significantly improve inference performance. Our experiments suggest that setting $\lambda_1$ and $\lambda_2$ to values around 0.5 often yields better results in such cases, as shown in Table~\ref{tab:task distribution gaps}. We hope this key insight helps readers better understand the practical role of activation derivatives and lays a solid foundation for future research in this direction.

\begin{table}[htbp]
\centering

\resizebox{1\columnwidth}{!}{
\begin{threeparttable}
\begin{small}
\renewcommand{\multirowsetup}{\centering}
\begin{tabular}{cccccc||cccccc}

\toprule
\multicolumn{6}{c||}{\textbf{OPT-125M}}                                         & \multicolumn{6}{c}{\textbf{LLaMA-2-7B}}                             \\ \hline
$\lambda_1$/$\lambda_2$ & 0       & 0.2           & 0.5             & 0.8     & 1       & $\lambda_1$/$\lambda_2$ & 0      & 0.2    & 0.5            & 0.8    & 1      \\ \hline
0     & 1007.98 & 992.46        & 969.23          & 983.60   & 970.02  & 0     & 98.28  & 101.39 & 102.03         & 100.34 & 100.20  \\
0.2   & 1007.04 & 1002.04       & \textbf{958.73} & 1010.90  & 981.99  & 0.2   & 99.23 & 101.27 & 95.17 & 97.99  & 102.43 \\
0.5   & 1074.76 & 1004.22       & 1002.85         & 969.67  & 1029.56 & 0.5   & 102.40  & 101.47 & 106.73         & 98.33  & 101.56 \\
0.8   & 1096.42 & 1070.81       & 1083.9          & 1046.11 & 1031.64 & 0.8   & 101.15 & 109.74 & 98.47          & 99.57  & 101.74 \\
1     & 1038.87 & 1069.44       & 983.91          & 1044.52 & 1027.64 & 1     & \textbf{92.68} & 103.81 & 99.23          & 103.27 & 96.47  \\ \hline
\multicolumn{6}{c||}{\textbf{LLaMA-2-13B}}                                      & \multicolumn{6}{c}{\textbf{LLaMA-3-8B}}                             \\ \hline
$\lambda_1$/$\lambda_2$ & 0       & 0.2           & 0.5             & 0.8     & 1       & $\lambda_1$/$\lambda_2$ & 0      & 0.2    & 0.5            & 0.8    & 1      \\ \hline
0     & 89.85   & 89.94         & 91.90            & 93.03   & 117.55  & 0     & 161.02 & 186.67 & 167.10          & 167.12 & 162.42 \\
0.2   & 88.52   & 87.84         & 85.20            & 88.89   & 103.4   & 0.2   & 141.99 & 162.85 & 162.91         & 179.72 & 152.58 \\
0.5   & 84.16   & 85.94         & 83.93           & 82.35   & 94.95   & 0.5   & \textbf{135.43} & 160.24 & 145.03         & 160.39 & 149.59 \\
0.8   & 89.70    & 80.80 & 84.00              & 85.12   & 91.66   & 0.8   & 160.77 & 153.90  & 159.70          & 158.45 & 160.97 \\
1     & \textbf{76.80}   & 86.04         & 88.72           & 86.04   & 90.61   & 1     & 151.47      & 149.73      & 147.86              & 146.94      & 145.70      \\ \bottomrule
\end{tabular}
\end{small}
\end{threeparttable}
}
\caption{Hyperparameters $\lambda_1$ and $\lambda_2$ search results of $D^2Prune$ for each model with 80\% sparsity and the corresponding perplexity on WikiText2; lower perplexity is better. }
\renewcommand{\arraystretch}{1.2}
\label{tab:models-r1-r2}
\end{table}

\begin{table}[t]
\centering

\resizebox{1.0\columnwidth}{!}{
\begin{threeparttable}
\begin{small}
\renewcommand{\multirowsetup}{\centering}
\begin{tabular}{cccccc}
\toprule
\multirow{2}{*}{Sparsity} & \multirow{2}{*}{Method}     & \multicolumn{2}{c}{Hyperparameter} & \multicolumn{2}{c}{Calibration data: C4} \\ \cline{3-6} 
                          &                             & $\lambda_1$        & $\lambda_2$       & HellaSwag: acc      & WikiText2: ppl     \\ \hline
\multirow{4}{*}{50}       & SparseGPT                   & -                & -               & 70.51               & 6.52               \\
                          & Wanda                       & -                & -               & 70.70               & 6.44               \\ \cline{2-6} 
                          & \multirow{2}{*}{$D^2Prune$} & 1                & 0               & 70.53               & \textbf{6.36}      \\
                          &                             & 0.5              & 0.5             & \textbf{70.78}      & 6.38               \\ \hline
\multirow{4}{*}{60}       & SparseGPT                   & -                & -               & 61.84               & 9.56               \\
                          & Wanda                       & -                & -               & 58.51               & 9.89               \\ \cline{2-6} 
                          & \multirow{2}{*}{$D^2Prune$} & 1                & 0               & 61.60               & \textbf{9.05}               \\
                          &                             & 0.5              & 0.5             & \textbf{62.05}      & 9.11               \\ \hline
\multirow{4}{*}{70}       & SparseGPT                   & -                & -               & 40.16               & 29.62              \\
                          & Wanda                       & -                & -               & 29.93               & 84.92              \\ \cline{2-6} 
                          & \multirow{2}{*}{$D^2Prune$} & 1                & 0               & 40.07               & \textbf{21.10}              \\
                          &                             & 0.5              & 0.5             & \textbf{45.81}      & 21.48              \\ \hline
\multirow{4}{*}{80}       & SparseGPT                   & -                & -               & 29.60               & 102.43             \\
                          & Wanda                       & -                & -               & 26.63               & 5107.20            \\ \cline{2-6} 
                          & \multirow{2}{*}{$D^2Prune$} & 1                & 0               & 28.43               & \textbf{92.68}              \\
                          &                             & 0.5              & 0.5             & \textbf{29.81}      & 98.94              \\ \bottomrule
\end{tabular}
\end{small}
\end{threeparttable}
}
\caption{Effect of $\lambda_1$, $\lambda_2$ settings under varying task distribution gaps ($D^2Prune$ on LLaMA-2-7B).}
\label{tab:task distribution gaps}
\end{table}

\textbf{Scaling factor: $s$}. In the weight importance evaluation, we introduce a scaling factor $s$ (see Algorithm~\ref{alg:$D^2Prune$}) when computing the activation norm to balance the magnitudes of weights and activations. This serves a role similar to normalization and has been widely adopted across various domains, such as activation scaling in quantization (e.g., SmoothQuant~\cite{xiao2023smoothquant}, OmniQuant~\cite{shao2023omniquant}) and gradient scaling in pruning (e.g., Pruner-Zero~\cite{dong2024pruner}). To further investigate the physical meaning of the scaling factor $s$, we derive its theoretical value based on activation norms. Specifically, we estimate the scaling ratio between $x$ and $y$ by analyzing the relationship between the input activation norm $||x||$ and the output activation norm $||y||$. According to the method of moments and the law of large numbers, we have:
\begin{equation}
\mathbb{E} _{\boldsymbol{x}}\left[ \boldsymbol{x}^2 \right] \rightarrow \frac{1}{n}\sum_{i=1}^n{\boldsymbol{x}_{i}^{2}}=\frac{||\boldsymbol{x}||_{2}^{2}}{n}
\label{eq:the law of large numbers}
\end{equation}
where, $n$ is the number of samples. It is also the activation length $L$ in this paper.

As shown in eq.~\ref{eq:the law of large numbers}, the theoretical value of the scaling factor corresponds to the length of the activation vector, which in our case is the sequence length $L$ ($s=L$ can be approximated when $L$ is large). In practice, $L$ is often manually set during preprocessing of the calibration data for model input—e.g., 2048 or 4096—and ideally matches the maximum sequence length of pre-trained model (if have sufficient memory resources in your GPU). In such cases, we recommend setting $s = L$, which can be determined automatically without requiring hyperparameter tuning.

\begin{figure*}[t] 
    \centering
    \includegraphics[width=\textwidth]{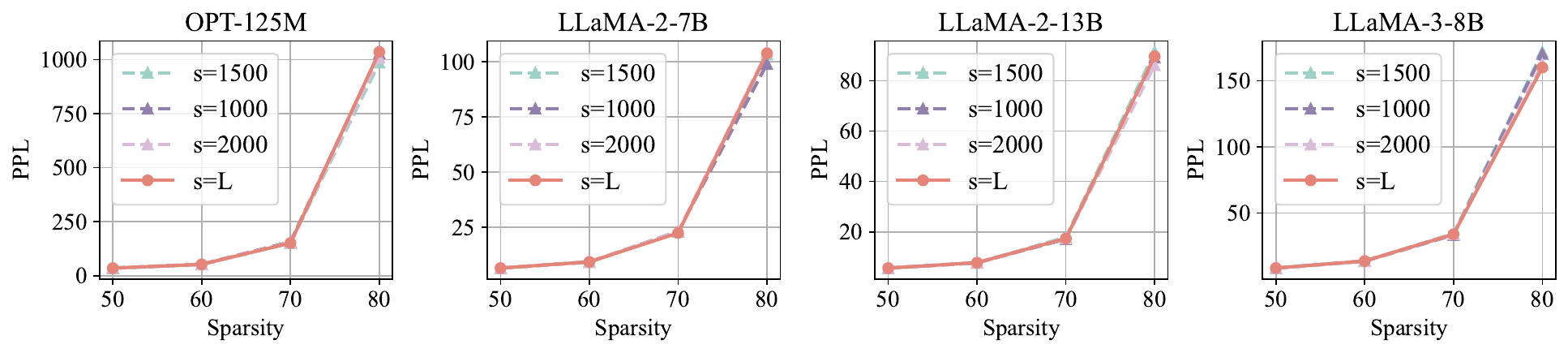} %
    \caption{Perplexity change of different pruning models with varying scaling factor $s$ on WikiText2} 
    \label{fig:s=L} %
\end{figure*}

\begin{figure*}[t] 
    \centering
    \includegraphics[width=\textwidth]{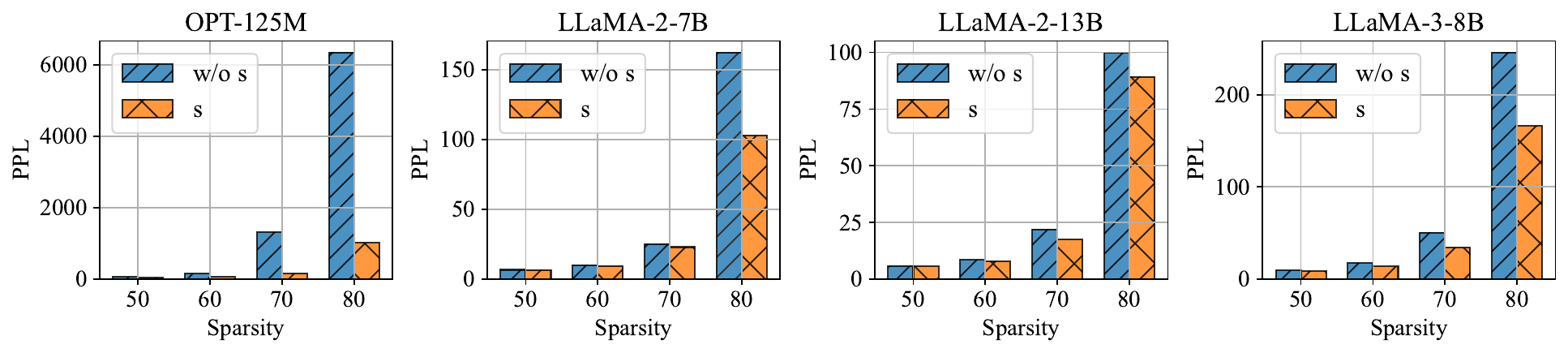} %
    \caption{Comparison of model perplexity on WikiText2 before and after introducing the scaling factor $s$ ("w/o s" denotes the setting without the scaling factor, while "s" indicates the setting with the scaling factor applied.} 
    \label{fig:s or not} %
\end{figure*}

\begin{figure*}
\subfloat[opt-125m]{
        \includegraphics[width=0.23\linewidth]{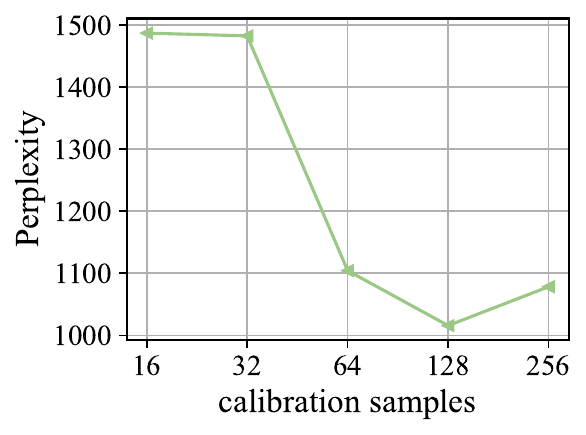}
        \label{fig:nsamples-opt-125m}
    }
\subfloat[llama-2-7b]{
        \includegraphics[width=0.23\linewidth]{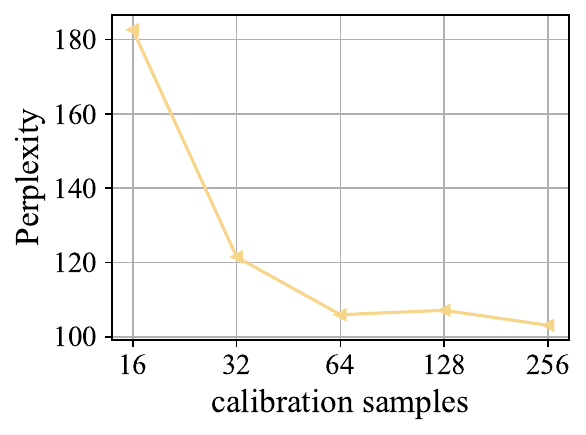}
        \label{fig:nsamples-plot-llama-2-7b}
    } 
\subfloat[llama-2-13b]{
        \includegraphics[width=0.23\linewidth]{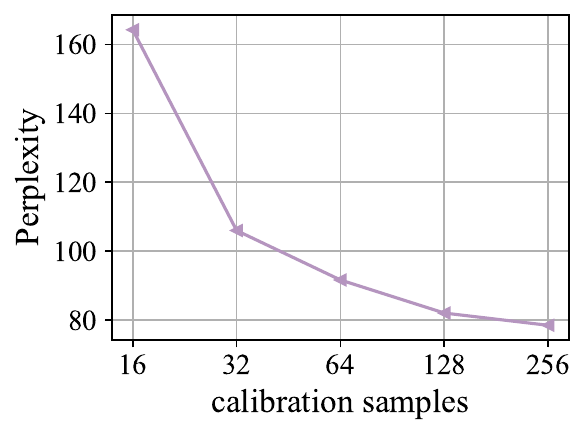}
        \label{fig:nsamples-plot-llama-2-13b}
    }
\subfloat[llama-3-8b]{
        \includegraphics[width=0.23\linewidth]{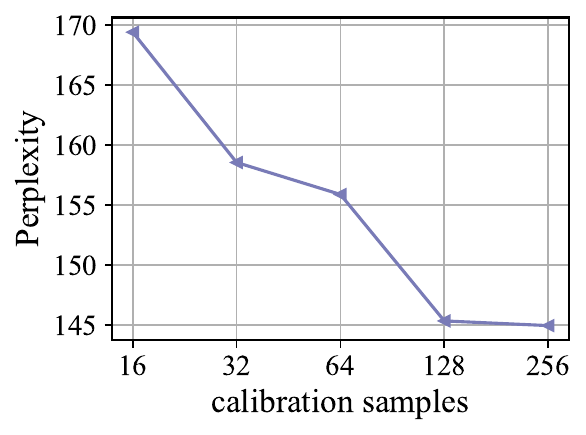}
        \label{fig:nsamples-plot-llama-3-8b}
    }
\caption{Perplexity sensitivity of different LLMs at 80\% sparsity for the calibration samples.}
\label{fig:calibration_samples_plot_llms}
\end{figure*}

Conversely, when $L$ is small, it is preferable to treat $s$ as a tunable hyperparameter, as it depends on the specific activation distribution. Based on our experiments, setting $s$ in the range of 1000 to 2000 typically yields better pruning performance, as illustrated in Figure~\ref{fig:s=L}. Moreover, as shown in Figure~\ref{fig:s or not}, pruning models with the scaling factor $s$ consistently outperforms those without it, highlighting the necessity of incorporating the scaling factor.
\textbf{Calibration samples}. Figure~\ref{fig:calibration_samples_plot_llms} show the ablation study results on the calibration data sample of $D^2Prune$ under 80\% sparsity condition for all the LLMs, which demonstrates that the calibration data sample size of 128 is a reasonable setting.

\subsection{Local Pruning vs. Global Pruning}
\label{sec:Local Pruning vs. Global Pruning}
To further evaluate the effectiveness of $D^2Prune$, we compare the performance of local and global pruning methods on the OPT-125M and LLaMA-2-7B models under varying sparsity levels. As shown in Table\ref{tab:global_local}, $D^2Prune$, a local pruning method that combines a Dual Taylor Expansion–based error estimator with attention distribution–aware dynamic weight update strategy, consistently outperforms global pruning baselines like SparseLLM\cite{bai2024sparsellm} (SparseLLM is a recent global pruning framework that decomposes the global optimization objective into coordinated subproblems using auxiliary variables and closed-form solvers for MLP structure of Transformers). While it achieves strong performance, particularly in high-sparsity regimes, $D^2Prune$ delivers superior results by minimizing local approximation errors more precisely and efficiently.

\begin{table}[htbp]
\centering
\resizebox{1\columnwidth}{!}{
\begin{threeparttable}
\begin{small}
\renewcommand{\multirowsetup}{\centering}
\begin{tabular}{cccccc}
\toprule
\multirow{2}{*}{Model}       & \multirow{2}{*}{Method} & \multicolumn{4}{c}{Sparsity}     \\ \cline{3-6} 
                             &                         & 50    & 60    & 70     & 80      \\ \hline
\multirow{3}{*}{OPT-125M} & SparseGPT               & 36.85 & 59.46 & 218.29 & 2140.55 \\
                             & SparseLLM               & 36.11 & 56.64 & 172.85 & 1778.7  \\
                             & $D^2Prune$                 & \textbf{34.98} & \textbf{52.1}  & \textbf{160.81} & \textbf{1038.87} \\ \hline
\multirow{3}{*}{LLaMA-2-7B}                   & SparseGPT               & 7.02  & 10.23 & 26.78  & 108.11  \\
                             & SparseLLM               & 7.06  & 10.26 & 28.87  & 118.22  \\
                             & $D^2Prune$                 & \textbf{6.96}  & \textbf{9.05}  & \textbf{21.1}   & \textbf{92.68}   \\  \bottomrule
\end{tabular}
\end{small}
\end{threeparttable}
}
\caption{Perplexity of Global Pruning vs. Local Pruning for OPT-125M and LLaMA-2-7B on WikiText2. the lower the perplexity, the better. SparseGPT and $D^2Prune$ are local pruning, while SparseLLM is global pruning. Our proposed pruning method $D^2Prune$ outperforms SparseGPT and SparseLLM.}
\label{tab:global_local}
\end{table}

\section{Theoretical Analysis and Verification of Dual Taylor Expansion on Error}
 \label{dual taylor expansion}
\begin{figure*}[t]
\centering
\subfloat[OPT-125M]{
        \includegraphics[width=0.23\linewidth]{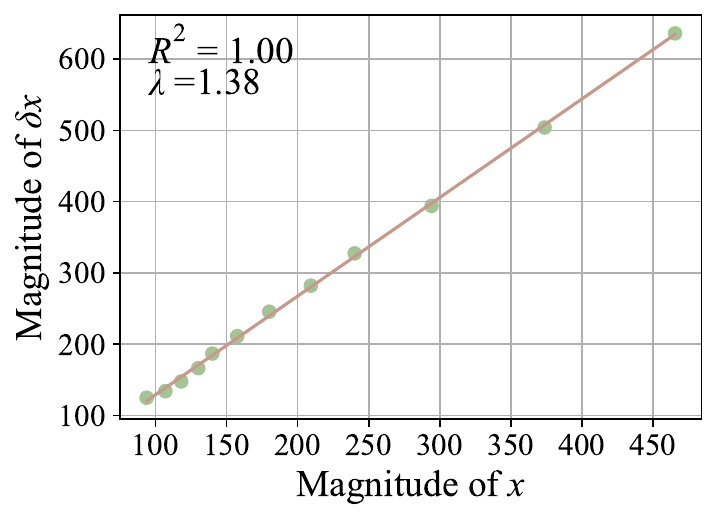}
        \label{fig:opt-125m-deltax_x-vs-x}
    }
\subfloat[LLaMA-2-7B]{
        \includegraphics[width=0.23\linewidth]{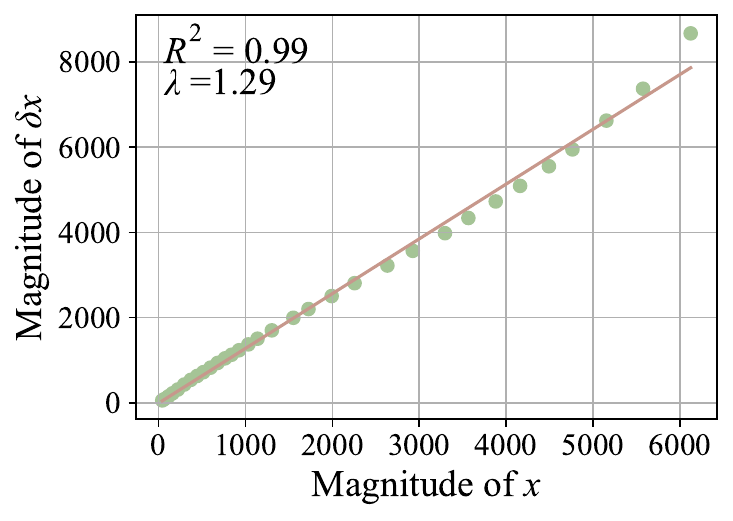}
        \label{fig:llama-2-7b-deltax_x-vs-x}
    }
\subfloat[LLaMA-2-13B]{
        \includegraphics[width=0.23\linewidth]{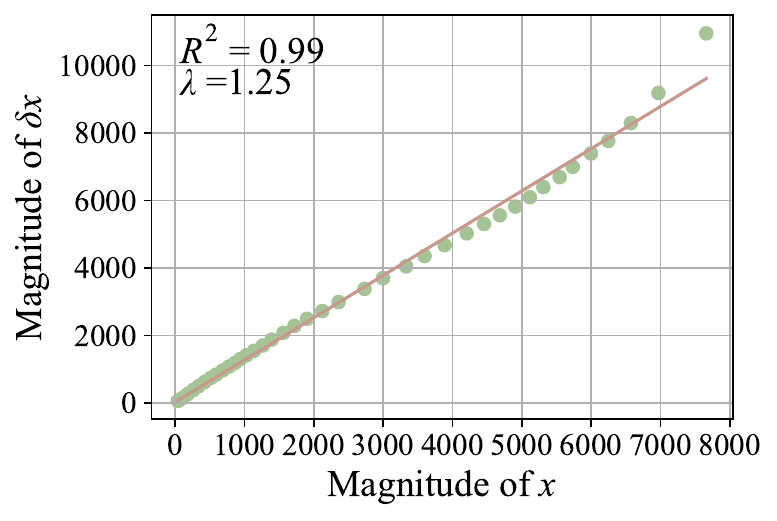}
        \label{fig:llama-2-13b-deltax_x-vs-x}
    }
\subfloat[LLaMA-3-8B]{
        \includegraphics[width=0.23\linewidth]{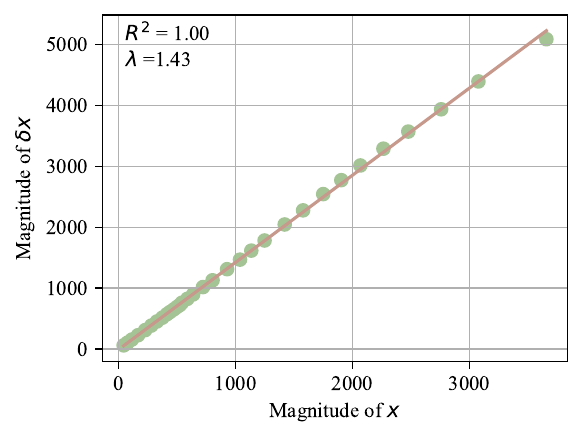}
        \label{fig:llama-3-8b-deltax_x-vs-x}
    }
\vspace{-2mm}
\caption{Layer-wise activation magnitude change relationship under different calibration data (C4-WikiText2). $R^2$ indicates the goodness of fit, the closer to 1, the better the fit is.}
\label{fig:c4-wiki-lambda}
\vspace{-3mm}
\end{figure*}

\begin{figure*}[t]
\centering
\subfloat[OPT-125M]{
        \includegraphics[width=0.23\linewidth]{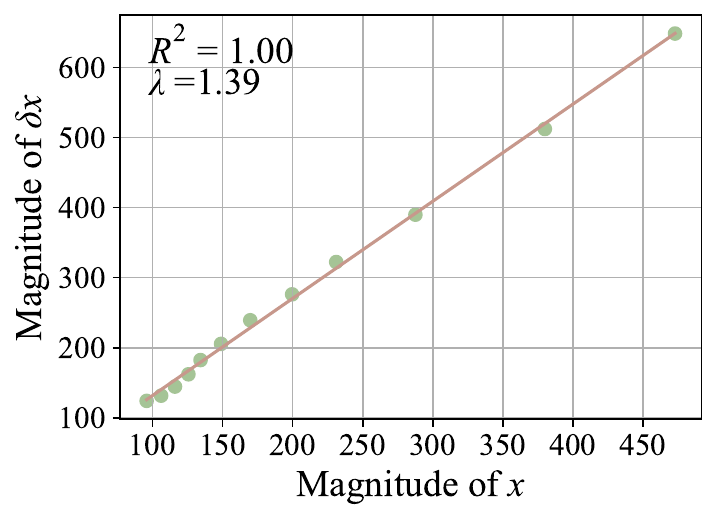}
        \label{fig:opt-125m-deltax_x-vs-x}
    }
\subfloat[LLaMA-2-7B]{
        \includegraphics[width=0.23\linewidth]{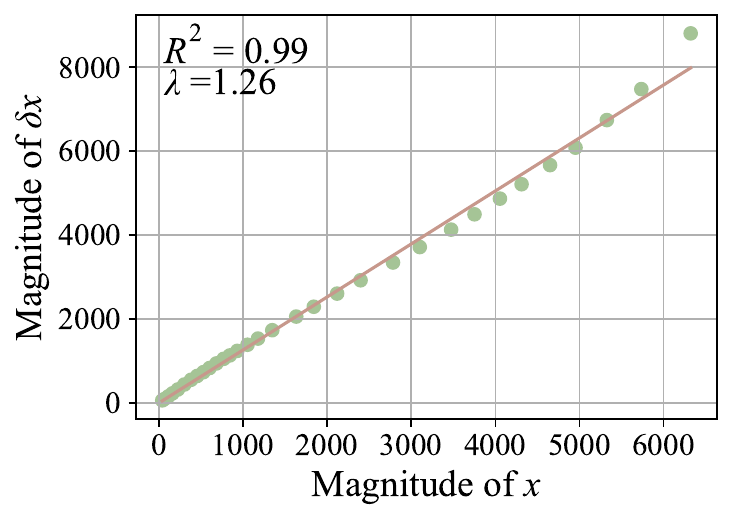}
        \label{fig:llama-2-7b-deltax_x-vs-x}
    }
\subfloat[LLaMA-2-13B]{
        \includegraphics[width=0.23\linewidth]{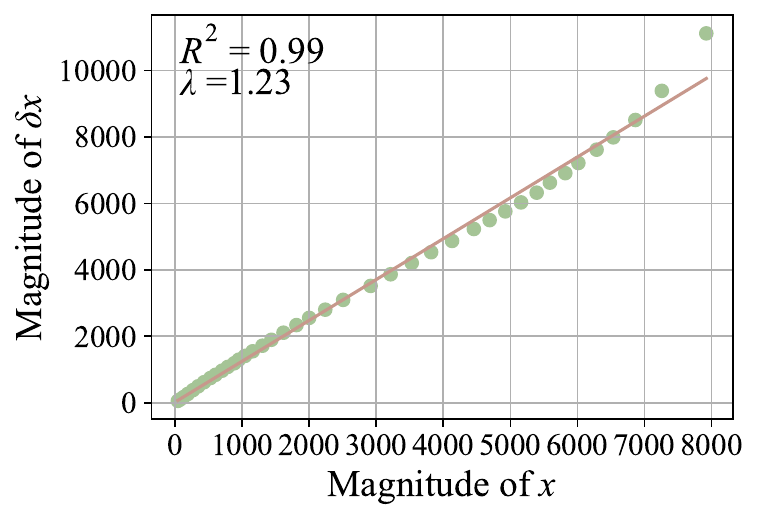}
        \label{fig:llama-2-13b-deltax_x-vs-x}
    }
\subfloat[LLaMA-3-8B]{
        \includegraphics[width=0.23\linewidth]{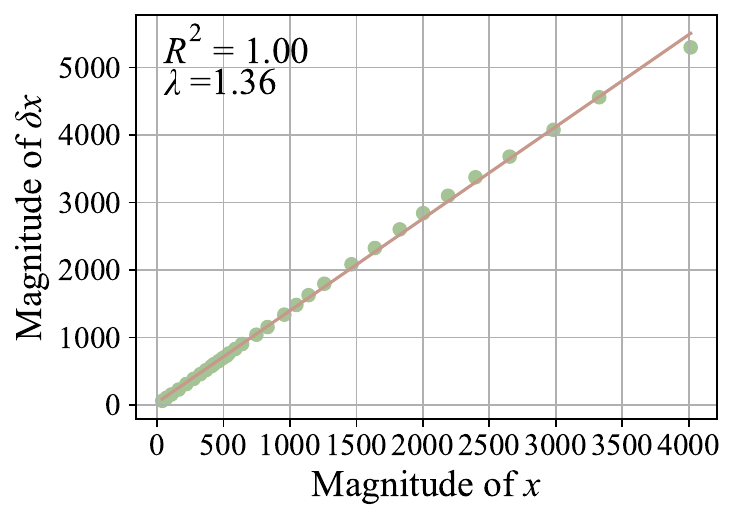}
        \label{fig:llama-3-8b-deltax_x-vs-x}
    }
\vspace{-2mm}
\caption{Layer-wise activation magnitude change relationship under different calibration data (C4-HellaSwag). $R^2$ indicates the goodness of fit, the closer to 1, the better the fit is.}
\label{fig:c4-hellaswag-lambda}
\vspace{-3mm}
\end{figure*}

\begin{figure*}[t]
\centering
\subfloat[OPT-125M]{
        \includegraphics[width=0.23\linewidth]{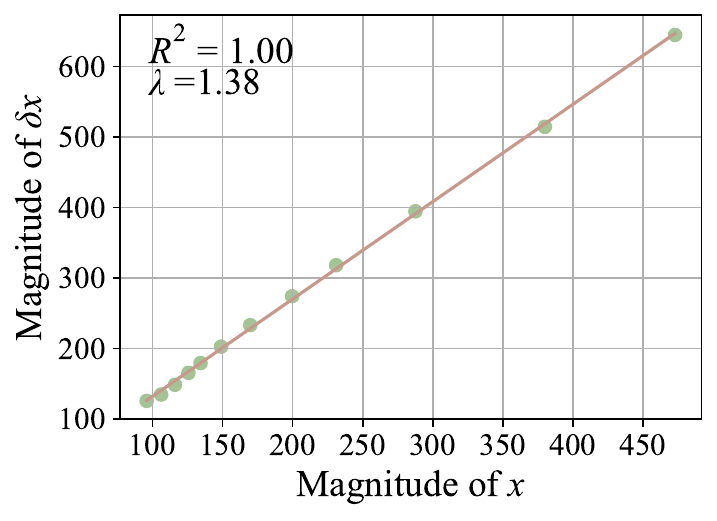}
        \label{fig:opt-125m-deltax_x-vs-x}
    }
\subfloat[LLaMA-2-7B]{
        \includegraphics[width=0.23\linewidth]{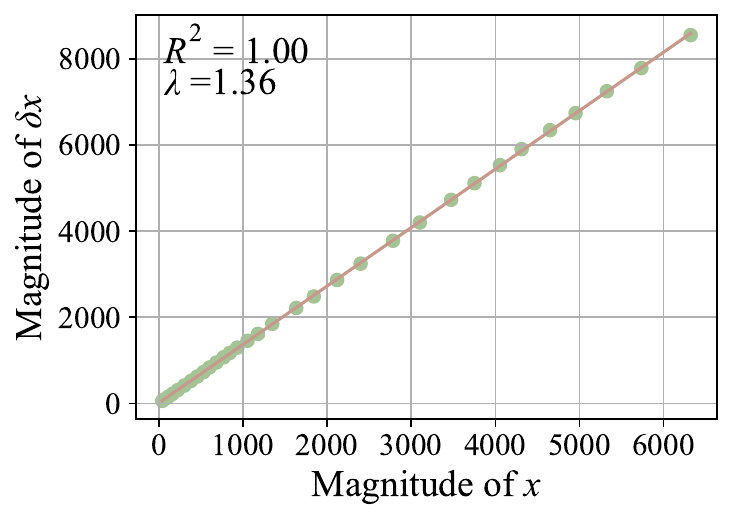}
        \label{fig:llama-2-7b-deltax_x-vs-x}
    }
\subfloat[LLaMA-2-13B]{
        \includegraphics[width=0.23\linewidth]{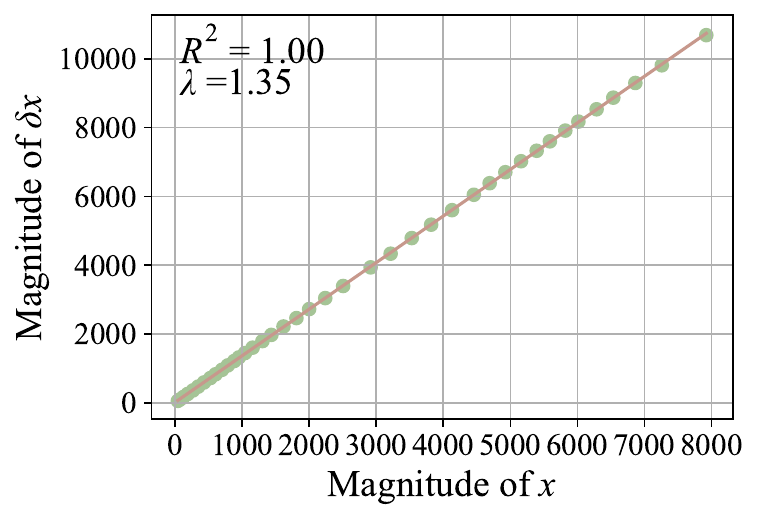}
        \label{fig:llama-2-13b-deltax_x-vs-x}
    }
\subfloat[LLaMA-3-8B]{
        \includegraphics[width=0.23\linewidth]{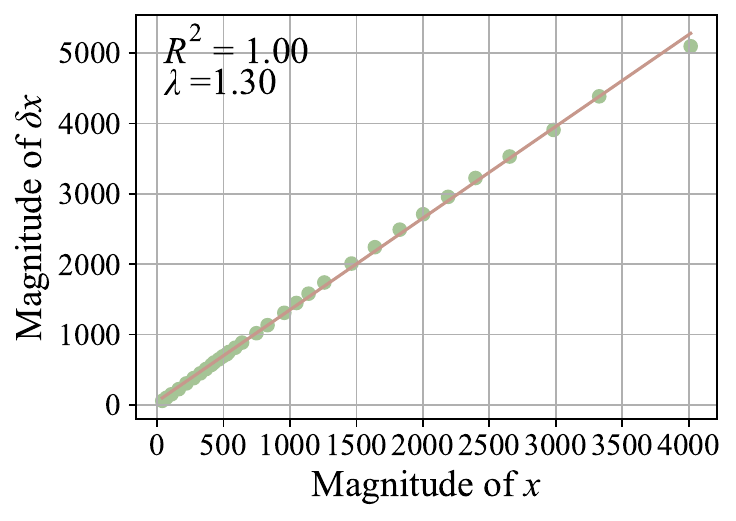}
        \label{fig:llama-3-8b-deltax_x-vs-x}
    }
\vspace{-2mm}
\caption{Layer-wise activation magnitude change relationship under different calibration data (WikiText2-HellaSwag). $R^2$ indicates the goodness of fit, the closer to 1, the better the fit is.}
\label{fig:wikitext-hellaswag-lambda}
\vspace{-3mm}
\end{figure*}

 \textbf{Assumption 1: $\delta \boldsymbol{x}=\lambda\boldsymbol{x}$} (in Eq. (8). As described in main text), 
 due to the distributional difference between the calibration data and the original batch training data
 of pre-trained LLM, we cannot ignore the impact of activation variations on error. Specifically, consider a pre-trained large language model where the original input activation of a certain layer is $\boldsymbol{x}_0$, the weight is $\boldsymbol{w}_0$, and the output is $
y={\boldsymbol{x}_0}^T\boldsymbol{w}_0$. Now, suppose we remove a certain weight such that
the error is $
E=||\boldsymbol{x}^T\boldsymbol{w}-{\boldsymbol{x}_0}^T\boldsymbol{w}_0||_{2}^{2}
$, where $\boldsymbol{w}=\boldsymbol{w}_0+\delta \boldsymbol{w}$ , and $
\boldsymbol{x}=\boldsymbol{x}_c=\boldsymbol{x}_0+\delta \boldsymbol{x} 
$ is the input activation when using calibration data. Therefore, $\delta \boldsymbol{x}=\boldsymbol{x}-\boldsymbol{x}_0=\boldsymbol{x}_c-\boldsymbol{x}_0$. 
Relative to the original batch training data, when pruning with the calibration data, the activation change is determined by the difference in the distributions of both. However, the original training data is unknown but fixed, so the relationship of the difference of between is mainly determined by the calibration data. Ideally, $\boldsymbol{\delta x}=\boldsymbol{x}-\boldsymbol{x}_0=\boldsymbol{x} + c$ (c is a constant) , exhibiting a linear relationship. Without loss of generality, assume that $\boldsymbol{\delta x}={\lambda} \boldsymbol{x}$, $\lambda$ is close to 1. To further verify the conjecture, we plot the layer-wise change of activation magnitude relationship under different data. As shown in Figure~\ref{fig:c4-wiki-lambda}, ~\ref{fig:c4-hellaswag-lambda}, ~\ref{fig:wikitext-hellaswag-lambda}, we can see that when any two of the three different calibration datasets are used as model inputs, the layer-wise activation change $\delta x$ exhibits a linear relationship with the activation $x$, with the fitted coefficient $\lambda$ consistently close to 1. Theoretically, when $\lambda = 1$ (the default setting in our experiments), no additional hyperparameters are introduced in the partial derivative of the activation. To compensate for the perturbation caused by activation changes, we introduce the hyperparameters $\lambda_1$ and $\lambda_2$ in Eq. (10), which is a reasonable design choice. The experimental results in Table~\ref{tab:models-r1-r2} also confirm the validity of the hyperparameter settings. 

\textbf{Assumption 2: The effect of the Hessian matrix cross terms is weak}. In Eq. (7) (detailed in main text), we ignore the cross terms of the Hessian matrix, i.e., $
\frac{1}{2}\delta \boldsymbol{w}^TH_{12}\delta \boldsymbol{x}+\frac{1}{2}\delta \boldsymbol{x}^T\boldsymbol{H}_{21}\delta \boldsymbol{w}=\delta \boldsymbol{w}^TH_{12}\delta \boldsymbol{x}
$ (where $
\boldsymbol{H}_{12}=\frac{\partial ^2E}{\partial \boldsymbol{w}\partial \boldsymbol{x}}=\boldsymbol{wx}^T+\left( \boldsymbol{w}^T\boldsymbol{x}-y \right) \boldsymbol{I}
$, $\boldsymbol{I}$ is the unit matrix). Since the pre-trained model has $
\frac{\partial E}{\partial \boldsymbol{w}}=\left( \boldsymbol{w}^T\boldsymbol{x}-y \right) \boldsymbol{x}=0
$ at local minima, we have $
\boldsymbol{H}_{12}=\boldsymbol{wx}^T
$. Considering the presence of Layer Normalization in Transformers (i.e., $
\mathrm{LayerNorm}\left( \boldsymbol{x} \right) =\frac{\boldsymbol{x}-\mu}{\sigma}
$, $\mu$, $\sigma$ are the expectation and variance of the activation $\boldsymbol{x}$, respectively)
, the input activations of each layer are normalized during the forward propagation on the calibration data, then $
\mathbb{E} _{\boldsymbol{x}}\left[ \boldsymbol{x}^T \right] =0
$
. This leads to an expectation of 0 for the cross terms, i.e., $
\mathbb{E} _{\boldsymbol{x}}\left[ \boldsymbol{wx}^T \right] =\boldsymbol{w}\mathbb{E} _{\boldsymbol{x}}\left[ \boldsymbol{x}^T \right] =0
$
. Therefore, the effect of the cross term $
\boldsymbol{H}_{12}
$ is weak and negligible compared to the Hessian matrix diagonal elements $\boldsymbol{H}_{11}$ and $\boldsymbol{H}_{22}$
, which are higher-order infinitesimals. 
Even in the presence of unnormalized activations within the attention mechanism, we compensate for the errors introduced by non-zero cross terms of the Hessian matrix through dynamic adjustment of $q$, $k$, and $v$ update configuration. The experimental results in Tables 2 and 3 (detailed in main text) further support this key insight. Compared to other methods such as SparseGPT and Wanda, our approach achieves better pruning performance without requiring additional Hessian-based computations (Just calculating the product term between the input and output activations paradigm with the weight magnitude, see~\ref{alg:$D^2Prune$}), thereby validating the reasonableness of our assumption.

\section{Q, K, V Weights Update Configurations Search}
\label{q,k,v update seach}
In the process of optimizing neural network structures (see Figure 4), $D^2Prune$ identifies the Q, K, and V update configurations that preserve the long-tail distribution characteristic of attention mechanisms, as shown in Table \ref{tab:hybrid pruning in 70 sparsity}, Table \ref{tab:hybrid pruning in 60 sparsity} as well as Table \ref{tab:hybrid pruning in 50 sparsity}. 

Considering the cumulative effect of layer-wise reconstruction errors (also referenced in Eq. (12) (detailed in main text), we reduce the original Q/K/V configuration search space from $\sum_{k=0}^3{C_{3}^{k}}=8$ combinations to 3 representative ones, and evaluate the corresponding perplexities. Under the constraint of model simplification, selecting the configuration that yields the lowest perplexity not only helps maintain inference performance after pruning but also strikes a favorable balance between model complexity and predictive capability, demonstrating the effectiveness of this pruning strategy.

\textbf{Uniform Layer-wise Q, K, V Update Configuration Search}.
For unstructured pruning with layer-wise consistent Q, K, V update configurations search, only three full forward passes are needed to identify the optimal update strategy. Here, we report the optimal settings for each model as follows: OPT-125M, LLaMA-2-7B, LLaMA-2-13B, LLaMA-3-8B, and LLaMA-70B correspond to {k: non-update}, {v: non-update}, {k: non-update}, {v: non-update}, and {k: non-update}, respectively (detailed experiments results in 3 (detailed in main text), 
Table~\ref{tab:hybrid pruning}
Table~\ref{tab:hybrid pruning in 70 sparsity}, Table~\ref{tab:hybrid pruning in 60 sparsity}, Table~\ref{tab:hybrid pruning in 50 sparsity}). The configuration {k: non-update} indicates that, after pruning, only the q and v weights are updated, while the k weights remain fixed.

\begin{table}[htbp]
\centering
\resizebox{1\columnwidth}{!}{
\begin{threeparttable}
\renewcommand{\multirowsetup}{\centering}
\begin{tabular}{ccccccc}
\toprule
\textbf{Method} & \cellcolor[HTML]{FFFFFF}\textbf{Weight-Update Layers} & \textbf{OPT-125M} & \textbf{LLaMA-2-7B} & \textbf{LLaMA-2-13B} & \textbf{LLaMA-3-8B} & \textbf{LLaMA-2-70B} \\ \hline
SparseGPT & All Layers & 2140.55 & 102.43 & 99.14 & 178.01 & 25.86 \\
Wanda & None & 1920.63 & 5107.2 & 1384.4 & 2245.91 & 156.68 \\ \hline
 & All Layers & 2469.78 & 101.88 & 82.75 & 170.92 & 24.86 \\ 
 & None & 1790.44 & 3561.51 & 1001.00 & 1414.32 & 150.94 \\ 
 & w/o q & 2077.36 & 104.63 & 89.19 & 230.83 & 24.59 \\
 & w/o k & \textbf{1038.87} & 118.79 & \textbf{76.80} & 190.48 & 25.61 \\
 & w/o v & 2410.76 & \textbf{92.68} & 83.58 & \textbf{151.47} & \textbf{21.37} \\
\multirow{-6}{*}{$D^2Prune$} & Dynamic Update & \textbf{1038.87} & \textbf{92.68} & \textbf{76.80} & \textbf{151.47} & \textbf{21.37} \\ \bottomrule
\end{tabular}
\end{threeparttable}
}
\caption{Ablations study of dynamic weight update for attention modules (q, k, v) on WikiText2. (in perplexity, unstructured 80\% sparsity)}
\label{tab:hybrid pruning}
\end{table}
\begin{table}[t]
\centering
\renewcommand{\arraystretch}{0.85} 
\centering
\resizebox{1\columnwidth}{!}{
\begin{threeparttable}
\renewcommand{\multirowsetup}{\centering}
\setlength{\tabcolsep}{1.45pt}
\begin{tabular}{ccccccc}
\toprule
Method                      & Weight-Update                          & Layer & OPT-125M        & LLaMA-2-7B     & LLaMA-2-13B    & LLaMA-3-8B     \\ \hline
SparseGPT                   & \checkmark               & All   & 218.29          & 29.62          & 18.20          & 38.85          \\
Wanda                       & \xmark                  & All   & 347.42          & 84.92          & 44.86          & 114.30         \\ \hline
\multirow{5}{*}{$D^2Prune$} & \checkmark              & All   & 202.50          & 23.45          & 17.00          & 38.64          \\ \cline{2-7} 
                            & \multirow{4}{*}{\xmark} & All   & 332.98          & 81.62          & 41.32          & 106.58         \\ \cline{3-7} 
                            &                                        & q     & 198.46          & 25.16          & 17.32          & 37.14          \\
                            &                                        & k     & \textbf{153.97} & 24.04          & \textbf{16.51} & 39.02          \\
                            &                                        & v     & 203.60          & \textbf{21.10} & 17.19          & \textbf{33.37} \\ \bottomrule
\end{tabular}
\end{threeparttable}
}
\caption{Ablations study of dynamic weight update for attention modules (q, k, v) on WikiText2. (in perplexity, unstructured 70\% sparsity)}
\label{tab:hybrid pruning in 70 sparsity}
\end{table}

\begin{table}[t]
\centering

\renewcommand{\arraystretch}{0.85} 
\centering
\resizebox{1\columnwidth}{!}{
\begin{threeparttable}
\renewcommand{\multirowsetup}{\centering}
\setlength{\tabcolsep}{1.45pt}
\begin{tabular}{ccccccc}
\toprule
Method                      & Weight-Update                          & Layer & OPT-125M       & LLaMA-2-7B    & LLaMA-2-13B   & LLaMA-3-8B     \\ \hline
SparseGPT                   & \checkmark               & All   & 59.46          & 9.56          & 7.77          & 14.40          \\
Wanda                       & \xmark                  & All   & 74.39          & 9.89          & 7.87          & 22.80          \\ \hline
\multirow{5}{*}{$D^2Prune$} & \checkmark              & All   & 55.39          & 9.46          & 7.58          & 13.66          \\ \cline{2-7} 
                            & \multirow{4}{*}{\xmark} & All   & 71.55          & 9.75          & 7.66          & 22.54          \\ \cline{3-7} 
                            &                                        & q     & 57.22          & 9.60          & 7.64          & 14.27          \\
                            &                                        & k     & \textbf{52.10} & 9.55          & \textbf{7.49} & 14.36          \\
                            &                                        & v     & 54.21          & \textbf{9.05} & 7.55          & \textbf{13.44} \\ \bottomrule
\end{tabular}
\end{threeparttable}
}
\caption{Ablations study of dynamic weight update for attention modules (q, k, v) on WikiText2. (in perplexity, unstructured 60\% sparsity)}
\label{tab:hybrid pruning in 60 sparsity}
\end{table}

\begin{table}[t]
\centering

\resizebox{1\columnwidth}{!}{
\begin{threeparttable}
\renewcommand{\multirowsetup}{\centering}
\setlength{\tabcolsep}{1.45pt}
\begin{tabular}{ccccccc}
\toprule
Method                      & Weight-Update                          & Layer & OPT-125M       & LLaMA-2-7B    & LLaMA-2-13B   & LLaMA-3-8B    \\ \hline
SparseGPT                   & \checkmark               & All   & 36.85          & 6.52          & 5.63          & 8.56          \\
Wanda                       & \xmark                  & All   & 38.88          & 6.44          & 5.58          & 9.06          \\ \hline
\multirow{5}{*}{$D^2Prune$} & \checkmark              & All   & 36.03          & 6.46          & 5.58          & 8.48          \\ \cline{2-7} 
                            & \multirow{4}{*}{\xmark} & All   & 38.15          & 6.43          & 5.54          & 8.92          \\ \cline{3-7} 
                            &                                        & q     & 36.01          & 6.50          & 5.60          & 8.55          \\
                            &                                        & k     & \textbf{34.98} & 6.49          & \textbf{5.53} & 8.57          \\
                            &                                        & v     & 36.25          & \textbf{6.36} & 5.56          & \textbf{8.34} \\ \bottomrule
\end{tabular}
\end{threeparttable}
}
\caption{Ablations study of dynamic weight update for attention modules (q, k, v) on WikiText2. (in perplexity, unstructured 50\% sparsity)}
\label{tab:hybrid pruning in 50 sparsity}
\end{table}

\textbf{Non-Uniform Layer-wise Q, K, V Update Configuration Search}. In the case of non-uniform Q, K, V update configuration search, each layer has three possible choices, leading to a total search space of $3^N$  for a model with $N$ layers. This exponential complexity makes exhaustive search impractical. To mitigate this, we conduct a theoretical analysis of the relationship between Q/K/V update strategies and attention distribution error (detailed in Section "Attention Distribution-Aware Dynamic Weight
 Update Strategy"). A key insight is that preserving critical weights within Q, K, and V plays a vital role in maintaining the attention distribution. From this perspective, whether a Q/K/V matrix should be updated is highly related to its corresponding outlier distribution. In theory, the higher the weight outlier ratio, the more likely it contains essential weights. Updating such weights may severely disrupt the attention structure, and therefore, these components should be handled with greater caution. Let $P$ denote the update probability of a given Q, K, or V weight matrix, and let $r$ represent its corresponding outlier ratio.  Based on our earlier analysis, in order to preserve critical weights, the update probability should be inversely related to the proportion of outliers. This relationship can be expressed as:
\begin{equation}
P\propto \left( 1-r \right). 
\end{equation}
This relationship reflects an intuitive principle: when Q, K, or V weight matrix contains a higher proportion of outliers (i.e., larger $r$),  its update probability should be correspondingly reduced to avoid disrupting the critical information encoded in the attention mechanism.
To automatically determine the Q, K, V update configuration for each layer, inspired by OWL, we adopt a dual Taylor expansion-based error estimation (see Eq. (12) in main text) to compute the outlier score for Q/K/V weights.
Specifically, we identify outliers as elements whose magnitudes is greater than the average value in each
layer. For a certain weight $
W\in \mathbb{R} ^{m\times n}
$ of Q/K/V, The outlier ratio $r$ is defined as:
\begin{equation}
\label{eq:outliers ratio}
r=\frac{|\left\{ w_{ij}\in W \right\} ||w_{ij}|>\mu |}{m\cdot n},    
\end{equation}
where $
w_{ij}
$ is the $(i,j)$-th element of the weight matrix $W$; $
\mu =\frac{\sum_{i=1}^m{\sum_{j=1}^n{|w_{ij}|}}}{m\cdot n}
$ is the mean of the absolute values of all elements in $W$. The numerator counts the number of elements whose magnitudes exceed $
\mu 
$
, while the denominator is the total number of elements in the matrix $W$.
To determine the update configuration for Q, K, and V in each layer, we first compute their respective outlier ratios by Eq.~\ref{eq:outliers ratio}, denoted as $r_Q$, $r_K$, and $r_V$. Let $
u\left( W \right) \in \left\{ 0,1 \right\} 
$ be an update indicator function for a weight matrix $W$, where $
u\left( W \right) =0
$ indicates that $W$ is not updated, and $
u\left( W \right) =1
$ indicates that $W$ is updated. We select the weight with the highest outlier ratio to remain unchanged, based on the intuition that preserving high-outlier weights helps maintain the attention distribution. Formally, the update rule is defined as:
\begin{equation}
\label{eq:qkv update function}
 u\left( W_i \right) =\begin{cases}
	0,\,\,\text{if } i=\underset{j\in \left\{ Q,K,V \right\}}{\mathrm{arg}\max}\,\,r_j\\
	1,\,\,\mathrm{otherwise}\\
\end{cases},
\end{equation}
where $W_i\in \left\{ W_Q,W_K,W_V \right\}$.
This strategy ensures that the most "outlier-dense" projection among Q, K, and V is preserved to minimize attention distortion. 

We compute the layer-wise Q, K, and V update configuration based on Eq.~\ref{eq:qkv update function}, with experimental results on WikiText2 (in perplexity) summarized in Table~\ref{tab:non-uniform layer wise q/k/v update}. On one hand, both the uniform and outlier-based non-uniform Q, K, V update configuration achieve significantly lower perplexity than random assignment, validating the effectiveness and reliability of our method. On the other hand, the perplexity difference between the outlier-guided non-uniform and the uniform Q/K/V update configuration is surprisingly small.

\begin{figure}[t] 
    \centering
    \includegraphics[width=\linewidth]{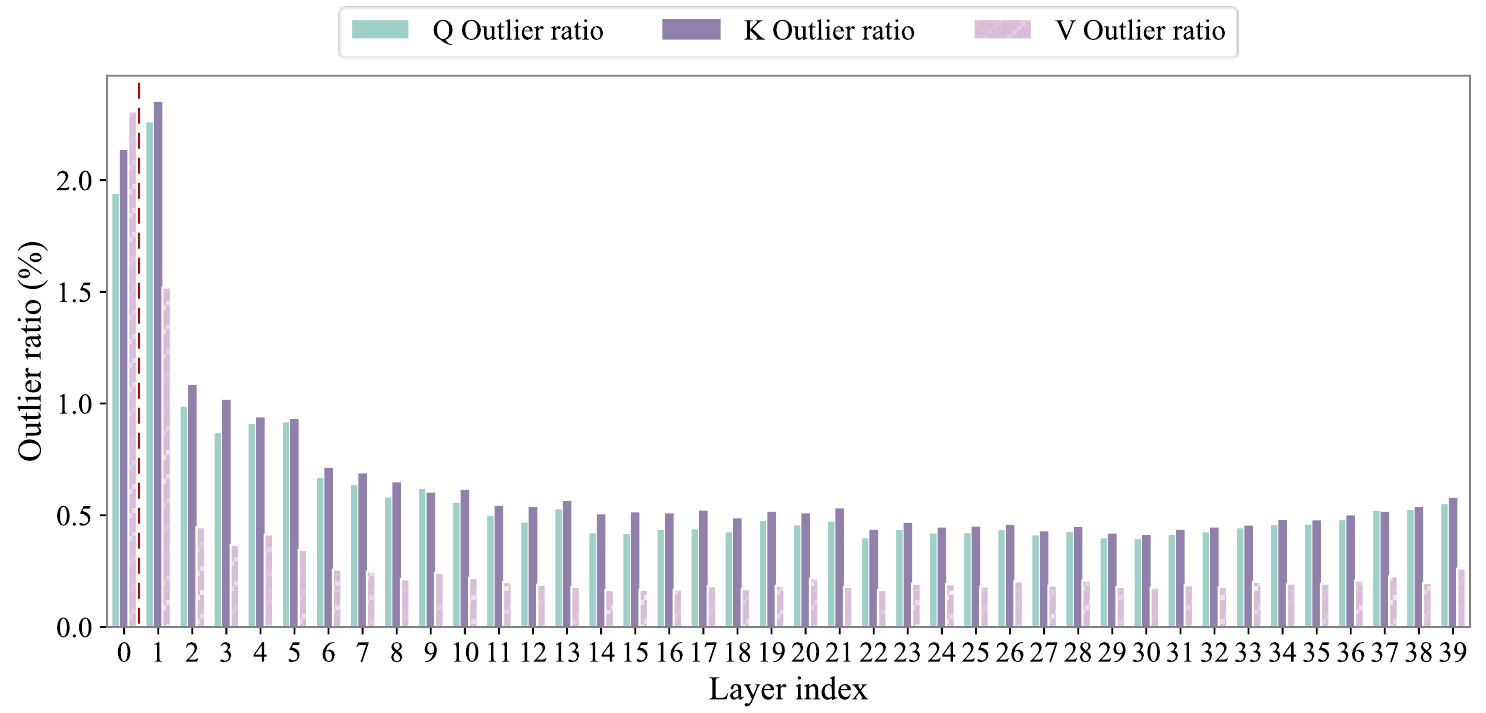} %
    \caption{Layer-wise distribution of Q, K, and V outlier ratios in LLaMA-2-13B} 
    \label{fig:llama-2-13b qkv outliers} %
\end{figure}

At first glance, this observation appears counterintuitive. In theory, large language models exhibit heterogeneous sensitivity to Q, K, and V across layers, and their corresponding outlier distributions should vary accordingly. However, our results indicate that such variations do not translate into meaningful differences in pruning performance. To further analyze this phenomenon, we visualize the layer-wise outlier ratios of Q, K, and V in LLaMA-2-13B, as shown in Figure~\ref{fig:llama-2-13b qkv outliers}. Interestingly, despite visible distributional differences across layers, the K weight matrix consistently shows the highest outlier ratio in almost all layers, except for the first layer, where the V weight matrix slightly dominates. This suggests that the model inherently favors a near-uniform update configuration across Q, K, and V.

We believe this novel finding sheds light on the internal structure of attention in large language models and offers valuable insights into how Q, K, and V representations are organized and utilized.
\begin{table}[t]
\centering

\resizebox{1\columnwidth}{!}{
\begin{threeparttable}
\begin{small}
\renewcommand{\multirowsetup}{\centering}
\begin{tabular}{cccccc}
\hline
Sparsity            & Layerwise Update & OPT-125M & LLaMA-2-7B & LLaMA-2-13B & LLaMA-3-8B \\ \hline
\multirow{3}{*}{50} & Uniform          & 34.98    & 6.36       & 5.53        & 8.34      \\
                    & Random           & 35.50    & 6.42       & 5.55        & 8.44      \\
                    & Non-Uniform      & 35.27    & 6.36       & 5.54        & 8.36      \\ \hline
\multirow{3}{*}{60} & Uniform          & 52.10    & 9.05       & 7.49        & 13.44     \\
                    & Random           & 54.13    & 9.28       & 7.42        & 41.08     \\
                    & Non-Uniform      & 52.34    & 9.05       & 7.42        & 13.67     \\ \hline
\multirow{3}{*}{70} & Uniform          & 160.81   & 21.10      & 16.51       & 33.37     \\
                    & Random           & 174.45   & 23.00      & 16.62       & 35.21     \\
                    & Non-Uniform      & 163.68   & 21.56      & 15.89       & 33.01     \\ \hline
\multirow{3}{*}{80} & Uniform          & 1038.87  & 92.68      & 76.80       & 151.47    \\
                    & Random           & 1382.89  & 107.19     & 80.43       & 187.50    \\
                    & Non-Uniform      & 1081.53  & 96.18      & 76.12       & 171.16    \\ 
\bottomrule
\end{tabular}
\end{small}
\end{threeparttable}
}
\caption{Comparison of uniform vs. non-uniform Layer-wise Q/K/V update vonfigurations across models at varying sparsity levels. “Random” denotes the random generation of Q/K/V updating configurations for each layer.}
\label{tab:non-uniform layer wise q/k/v update}
\end{table}

\section{Pruing Efficiency and Speedup}\label{sec:pruning_efficiency}
\textbf{Pruning efficiency}.
$D^2Prune$ has a lower theoretical computational complexity than SparseGPT.We compare the empirical pruning speed of various methods by measuring the total pruning time on an NVIDIA A40 GPU. The results are shown in Table~\ref{tab:prune_efficiency}. Notably, our proposed method achieves better pruning performance than weight-update approaches like SparseGPT without introducing additional computational overhead. Compared to non-weight-update methods (Wanda and Pruner-Zero), our approach attains significantly higher accuracy under high sparsity with minor overhead.

\textbf{Speedup}.
We evaluate the inference speedup enabled by semi-structured 2:4 sparsity on an NVIDIA A100 GPU. Following the evaluation protocols of SparseGPT~\cite{frantar2023sparsegpt} and Wanda~\cite{sun2023simple}, we measure the latency of matrix multiplications in the model’s linear layers. The simulation is performed using high-performance GEMM kernels from the NVIDIA Cutlass library. Table~\ref{tab:speedup} reports results on LLaMA-2-7B with batch size 1 and sequence length 
2048.
The results demonstrate that semi-structured 2:4 sparsity yields a significant inference speedup (approximately 1.3×) for the linear layers in large language models.
\begin{table}[htbp]
\centering
\small
\renewcommand{\arraystretch}{1.2}
\begin{tabular}{ccccc}
\toprule
\multirow{2}{*}{Method} & OPT & \multicolumn{2}{c}{LLaMA-2} & LLaMA-3 \\ \cmidrule(lr){2-5}
                        & 125M & 7B & 13B & 8B \\
\midrule
SparseGPT               & 49.13 & 963.68 & 1727.61 & 1893.83 \\
Wanda                   & 17.61 & 269.45 & 455.62  & 604.66  \\
Pruner-Zero             & 17.50 & 268.93 & 504.84  & 617.72  \\
$D^2Prune$              & 47.29 & 952.08 & 1617.51 & 1876.23 \\
\bottomrule
\end{tabular}
\caption{Comparison of time overhead used for pruning models (in seconds).}
\label{tab:prune_efficiency}
\end{table}

\begin{table}[htbp]
\centering
\small
\renewcommand{\arraystretch}{1.2}
\begin{tabular}{lccc}
\toprule
LLaMA Layer   & Dense & 2:4  & Speedup \\
\midrule
q/k/v/o\_proj & 2.38  & 1.79 & 1.33×   \\
up/gate\_proj & 3.29  & 2.43 & 1.35×   \\
down\_proj    & 1.66  & 1.25 & 1.32×   \\
\bottomrule
\end{tabular}
\caption{Speedup of matrix multiplication (ms) in LLaMA-2-7B, for semi-structured 2:4 sparsity.}
\label{tab:speedup}
\end{table}

\section{Limitations and Future Works}
\label{sec:Limitations and Future Works}
Although $D^2Prune$ optimizes the error minimization method in pruning, further improving the accuracy of pruning mask selection and weight updates, and making a significant step forward in the effective pruning for LLMs, attention-awareness for pruning mask selection and weight updates remains local suboptimal solutions. In our work, we still assume uniform sparsity, meaning that the pruning sparsity for each layer is the same and equal to the global sparsity. However, since different layers in LLMs have varying redundancy degree and pruning sensitivity, how to achieve an effective balance between sparsity and attention awareness, and use non-uniform sparse pruning for attention error compensation, is a significant direction. Finally, although we have explored post-training pruning compression techniques, we believe that considering instruction fine-tuning for different tasks and collaborative quantization compression will also be an important future research area.

\section{Related Work}
Post-training pruning has emerged as a critical technique for compressing Large Language Models (LLMs) \cite{kwon2022fast, zhang2024plug, reyhan2024novel}, offering the advantage of eliminating the need for retraining while significantly reducing computational resource requirements.
In this section, we briefly review post-training pruning techniques, categorizing them into the non-weight-update and weight-update pruning methods. 
Non-weight-update pruning defines a pruning metric to measure the importance of weights and then directly prunes unimportant weights without updating the remaining weights.
For instance, Wanda~\cite{sun2023simple} introduces an improved magnitude-based criterion that combines activations with weights. 
Building upon this, Pruner-Zero~\cite{dong2024pruner} employed symbolic regression and genetic programming to automate the discovery of pruning evaluation functions. 
%
On the other hand, weight-update pruning methods are generally based on (OBS)~\cite{hassibi1993optimal} for pruning mask selection and weight update, aiming to minimize layer-wise errors introduced by pruning.
Typically,
SparseGPT~\cite{frantar2023sparsegpt} utilizes Hessian matrices and calibration data to select masks and update weights. 

Recent advances have further improved weight-update pruning. The ADMM-Grad method~\cite{bovza2024fast} introduces an alternating direction method of multipliers (ADMM) for fast and effective weight updates after pruning, achieving state-of-the-art results across various LLMs. Meanwhile, LLM Surgeon~\cite{van2023llm} offers a general framework for unstructured, semi-structured, and structured pruning. It leverages Kronecker-factored curvature approximations to dynamically allocate pruning across layers and update weights efficiently, delivering top-tier performance in structured pruning of LLMs. SparseLLM~\cite{bai2024sparsellm} optimizes the pruning process of MLP modules in LLMs decoder layer through a global optimization approach.
\end{document}